%% file: main.tex
\PassOptionsToPackage{dvipsnames,table}{xcolor}
\ifdefined\pdfminorversion
\fi
\documentclass{article}
\usepackage{iclr2027_conference,times}
\input{math_commands.tex}

\usepackage{float}
\usepackage{hyperref}
\usepackage{xurl}
\usepackage{graphicx}
\usepackage{booktabs}
\usepackage{tabularx}
\usepackage{multirow}
\usepackage{makecell}
\usepackage{amsmath}
\usepackage{amssymb}
\usepackage{xcolor}
\usepackage{adjustbox}
\usepackage{enumitem}
\usepackage{placeins}
\usepackage{needspace}
\usepackage[breakable]{tcolorbox}
\hypersetup{colorlinks=true,linkcolor=OrangeRed,citecolor=MidnightBlue,urlcolor=OrangeRed}

\usepackage[capitalise,nameinlink,sort]{cleveref}
\crefname{figure}{Fig.}{Figs.}
\Crefname{figure}{Figure}{Figures}
\crefname{table}{Tab.}{Tabs.}
\Crefname{table}{Table}{Tables}
\creflabelformat{equation}{#2#1#3}

\definecolor{MindNavy}{HTML}{25364A}
\definecolor{MindDirect}{HTML}{3F78C5}
\definecolor{MindDirectPale}{HTML}{E9F1FC}
\definecolor{MindGrounded}{HTML}{8965BA}
\definecolor{MindGroundedPale}{HTML}{F2ECF9}
\definecolor{MindTeal}{HTML}{5D9E9A}
\definecolor{MindTealPale}{HTML}{E8F4F2}
\definecolor{MindAmber}{HTML}{D49938}
\definecolor{MindAmberPale}{HTML}{FFF5E1}
\definecolor{MindAdaptPale}{HTML}{EEF1F4}

\graphicspath{{figures/}}
\newcommand{\citeblue}[1]{\textcolor{MidnightBlue}{\citep{#1}}}

\title{MindView: Benchmarking Generative\\[-2pt] Multi-View Spatial Intelligence}

\iclrfinalcopy
\newcommand{\MindViewAuthorCell}[2]{%
  \makebox[0.24\textwidth][c]{\textbf{#1}\textsuperscript{#2}}}
\author{%
{\normalsize
\makebox[0.96\textwidth][c]{%
\MindViewAuthorCell{Yanbo Wang}{1,2}%
\MindViewAuthorCell{Siqi Luo}{3,2}%
\MindViewAuthorCell{Siyu Zhang}{4,2}%
\MindViewAuthorCell{Chenming Zhu}{5}}\\[3pt]
\makebox[0.96\textwidth][c]{%
\MindViewAuthorCell{Wenzhao Zheng}{6}%
\MindViewAuthorCell{Qian Yu}{2}%
\MindViewAuthorCell{Juncheng Yan}{1,2}%
\MindViewAuthorCell{Bin Fu}{2}}\\[3pt]
\makebox[0.96\textwidth][c]{%
\MindViewAuthorCell{Tai Wang}{2}%
\MindViewAuthorCell{Yu Qiao}{2}%
\MindViewAuthorCell{Yihao Liu}{2,\textdagger}%
\MindViewAuthorCell{Jiwen Lu}{1,\textdagger}}}\\[10pt]
{\small
\textsuperscript{1}Tsinghua University\qquad
\textsuperscript{2}Shanghai Artificial Intelligence Laboratory\\[3pt]
\textsuperscript{3}Shanghai Jiao Tong University\qquad
\textsuperscript{4}Fudan University\\[3pt]
\textsuperscript{5}The University of Hong Kong\qquad
\textsuperscript{6}University of California, Berkeley}\\[6pt]
{\small
\hypersetup{urlcolor=black}
\href{mailto:liuyihao@pjlab.org.cn}{\texttt{liuyihao@pjlab.org.cn}}\quad
\href{mailto:lujiwen@tsinghua.edu.cn}{\texttt{lujiwen@tsinghua.edu.cn}}}}

\makeatletter
\renewcommand{\@maketitle}{%
  \vbox{\hsize\textwidth
    {\raggedright\LARGE\scshape\@title\par}
    \vspace{14pt}
    {\centering\normalsize\@author\par}
    \vspace{12pt}}}
\makeatother
\hypersetup{
  pdftitle={MindView: Benchmarking Generative Multi-View Spatial Intelligence},
  pdfauthor={Yanbo Wang, Siqi Luo, Siyu Zhang, Chenming Zhu, Wenzhao Zheng, Qian Yu, Juncheng Yan, Bin Fu, Tai Wang, Yu Qiao, Yihao Liu, Jiwen Lu}
}

\begin{document}
\maketitle
% Explain the shared dagger in a first-page footnote; keep email addresses above.
% The repository link is an unmarked footer line, independent of the abstract.
\begingroup
\renewcommand{\thefootnote}{\fnsymbol{footnote}}
\footnotetext[2]{Corresponding authors.}
\footnotetext[0]{Project page: 
\href{https://yanbo-23.github.io/MindView/}{https://yanbo-23.github.io/MindView/}}
\endgroup

% Keep the main-text fonts, line spacing, and float anchors unchanged.
% Limit compact spacing to the main text.  Ragged bottoms prevent short pages
% from stretching the gaps between figures, tables, equations, and headings.
\begingroup
\raggedbottom
\setlength{\parskip}{4pt}
\setlength{\textfloatsep}{8pt minus 2pt}
\setlength{\floatsep}{6pt minus 2pt}
\setlength{\abovecaptionskip}{6pt}

\input{sections/abstract}

\input{sections/introduction}
\input{sections/related_work}
\input{sections/benchmark}
\input{sections/experiments}
\input{sections/conclusion}

% These unnumbered statements are outside the main-text page limit.
\clearpage
\endgroup
\input{sections/statements}

{\small
\bibliographystyle{iclr2027_conference}
\bibliography{references}
}

\clearpage
% The main-text group has restored all original spacing for the appendix.
\appendix
\input{sections/appendix}

\end{document}

%% file: math_commands.tex
\usepackage{amsmath,amsfonts,bm}

\def\eqref#1{equation~\ref{#1}}
\def\1{\bm{1}}

\DeclareMathAlphabet{\mathsfit}{\encodingdefault}{\sfdefault}{m}{sl}
\SetMathAlphabet{\mathsfit}{bold}{\encodingdefault}{\sfdefault}{bx}{n}

%% file: sections/abstract.tex
\begin{abstract}

    Spatial intelligence has recently been assessed through generative tasks such
    as spatially grounded image editing, yet language-guided generation of spatially
    consistent views from partial multi-view observations remains underexplored.\ %
    % Generating such views tests spatial intelligence through cross-view integration,
    % language-conditioned viewpoint inference, and unseen-view rendering.\ %
    % Related evaluation settings cover only parts of this chain: multi-reference
    % generation emphasizes content fidelity, whereas pose-conditioned novel-view
    % synthesis assumes a known target pose.\ %
    We introduce \textbf{MindView}, a benchmark for generative multi-view spatial
    intelligence with 1,500 samples from 513 scenes across three indoor datasets.
    Its seven tasks form two families: geometry-dominant tasks specify changes in
    camera position and orientation, while
    semantics-dominant tasks require inferring target viewpoints by grounding
    referred objects and interpreting spatial relations and reference-frame cues.\ %
    Evaluation of eleven generators reveals substantial room for improvement on
    MindView.\ %
    A paired context analysis further shows that all eleven generators obtain lower
    overall scores under the expanded context setting.\ %
    These findings motivate us to investigate whether task-specific supervision can
    improve target-view generation and whether models can use additional multi-view
    evidence effectively.\ %
    We therefore fine-tune BAGEL on the MindView training dataset, 
    improving its overall score by 24.3 points and yielding gains across all seven MindView tasks.
    Beyond these generation gains, the fine-tuned model also yields higher scores on three 
    spatial-understanding benchmarks without training on their annotations,
    suggesting that learning to generate spatially grounded views may benefit
    spatial understanding.\ %
    To improve evidence utilization under expanded context, we propose
    \textbf{MindCue}, a training-free spatial evidence selector guided
    by a target-view hypothesis. MindCue reverses the drop in overall score under expanded context
    and outperforms the standard-context baseline.
    Overall, MindView provides a concrete testbed for generative multi-view spatial
    intelligence.\ %
    Our findings further highlight task learning and query-guided evidence selection
    as complementary strategies for improving performance in this setting.\ %
\end{abstract}

%% file: sections/introduction.tex
\section{Introduction}

\begin{figure*}[!t]
    \centering
    \includegraphics[width=\textwidth]{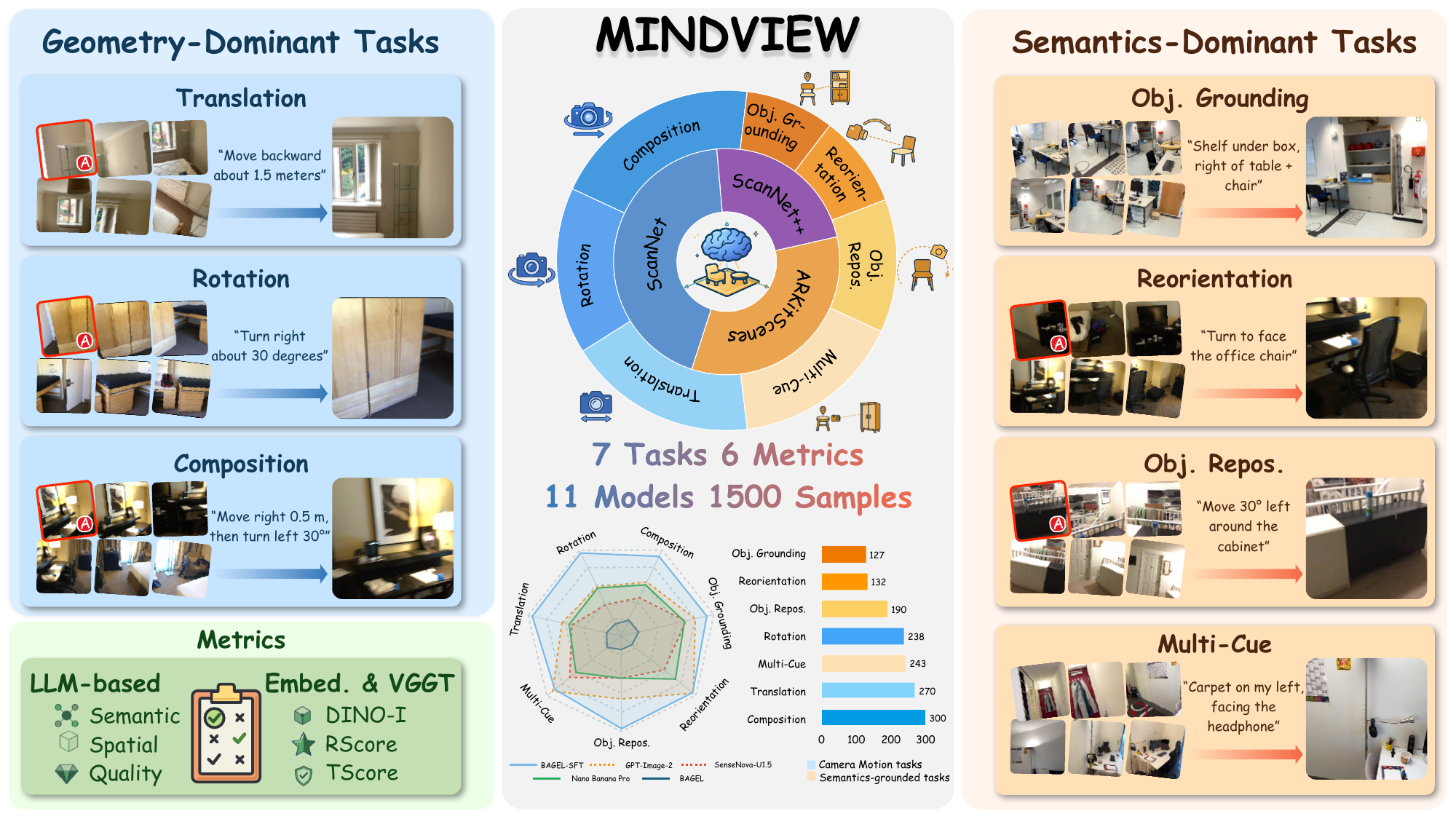}
    \caption{Overview of MindView, showing seven task examples, benchmark statistics, six metrics, and representative results. Red \textbf{A} marks anchor views, and targets are real images.}
    \label{fig:teaser}
\end{figure*}

Benchmarks such as MindCube \citeblue{exp_mindcube2025}, MMSI-Bench \citeblue{exp_mmsi2026}, and SAT-v2 \citeblue{exp_sat2025} evaluate the spatial intelligence of multimodal understanding models through question answering.\ %
On the generation side, GenSpace \citeblue{genspace2025} and GSI-Bench \citeblue{gsibench2026} have begun this exploration by asking whether spatial instructions are realized in a generated or edited image.\ %
We focus on evaluating \textit{generative multi-view spatial intelligence}
through language-guided target-view generation from partial observations
of the same scene (\Cref{fig:teaser}).\ %
This task requires models to integrate evidence across views,
infer the requested viewpoint, and predict what would be visible
from it while remaining consistent with the observed scene.\ %

Related work differs in how reference images are used and how target viewpoints are specified.\ %
Multi-reference benchmarks such as MICE-Bench \citeblue{micebench2026} and MultiBanana \mbox{\citeblue{multibanana2026}} emphasize preserving and combining reference content, which alone does not establish that a generated image depicts the requested viewpoint of the observed scene.\ %
Pose-conditioned methods such as ZeroNVS \citeblue{zeronvs2024} evaluate view synthesis with the target camera supplied, leaving language-conditioned viewpoint inference outside their task setting.\ %
More recently, Astra \citeblue{astra2026} uses language-controlled novel-view generation
to support agentic spatial question answering and evaluates generated
views against real targets for pose and content consistency.\ %
However, its generation evaluation focuses on explicit camera motions
with a single dominant degree of freedom.\ %
We seek a unified evaluation of individual and composed
camera motions and target viewpoints inferred from object
references, spatial relations, and reference-frame cues.\ %

We introduce \textbf{MindView}, a benchmark of \textbf{1,500 samples} from \textbf{513 scenes} in three indoor datasets (\Cref{fig:teaser}).\ %
Its seven tasks form two families: \textit{Geometry-Dominant} tasks specify camera translations, rotations, or their ordered composition, while \textit{Semantics-Dominant} tasks require viewpoint inference from referred objects, spatial relations, and reference-frame cues.\ %
The model receives only RGB observations and a language query, with the target image and numerical camera parameters withheld.\ %
Six complementary metrics assess content fidelity, spatial agreement, rendering quality, and geometric consistency to identify content and spatial errors an aggregate score may obscure.\ %

% Evaluation of eleven generators reveals recurring difficulties with composed camera motions and viewpoints specified by multiple scene cues.\ %
% A paired analysis on the same samples further shows that all eleven obtain lower Overall scores when context expands from six to ten views, despite greater coverage of target-view content.\ %
% These observations motivate two questions: can task-specific supervision improve target-view generation, and can query-guided selection make additional observations useful?\ %

Across eleven off-the-shelf generators, preserving scene content
while matching the requested viewpoint remains challenging
(\Cref{tab:final-c6-overall}).\ %
Expanding context from six to ten views increases target-view coverage from 88.4\% to 92.0\%, yet lowers Overall for all eleven generators (\Cref{fig:context-budget}).\ %
These findings raise two questions: \textit{can task-specific supervision improve target-view generation, and can query-guided evidence selection help models benefit from additional observations?}\ %

To address these questions, we curate a target-view generation training set and fine-tune BAGEL \citeblue{exp_bagel2025}, obtaining BAGEL-SFT with a 24.3-point Overall gain and improvements across all seven tasks.\ %qi shi
BAGEL-SFT also scores higher on three spatial-understanding benchmarks without training on their annotations, providing preliminary evidence of transfer from generation training to spatial understanding.\ %
To improve evidence utilization, we propose \textbf{MindCue}, a training-free spatial evidence selector using a target-view hypothesis to select supplementary observations from the expanded context.\ %
On both BAGEL and BAGEL-SFT, MindCue reverses the expanded-context degradation and outperforms the standard-context baseline and count-matched random selection.\ %

Our main contributions are summarized as follows:

\begin{enumerate}[leftmargin=*,itemsep=1pt,topsep=2pt]
    \item We introduce \textbf{MindView}, a benchmark for generative multi-view spatial intelligence, revealing two challenges across eleven generators: limited content fidelity and spatial consistency, and performance degradation with excessive context.\ %
    \item We curate target-view generation training data and show that task-specific supervision improves BAGEL across all seven tasks, with preliminary evidence of transfer to spatial understanding.\ %
    \item We propose \textbf{MindCue}, a training-free, query-guided evidence selector that improves both BAGEL and BAGEL-SFT over fixed and count-matched random baselines.
\end{enumerate}

%% file: sections/related_work.tex
\section{Related Work}

\textbf{Spatial Intelligence Benchmarks and Generative Reasoning.}
Spatial intelligence is evaluated through both question answering and image generation.\ %
MindCube \citeblue{exp_mindcube2025}, MMSI-Bench \citeblue{exp_mmsi2026}, and SAT-v2 \citeblue{exp_sat2025} assess spatial understanding through question answering, whereas GenSpace \citeblue{genspace2025} and GSI-Bench \citeblue{gsibench2026} evaluate whether generated images satisfy spatial instructions.\ %
The latter also reports understanding gains from generation-focused training.\ %
Thinking with Novel Views \citeblue{twnv2026} uses generated views for spatial question answering, while UniG2U-Bench \citeblue{unig2u2026} examines unified models' understanding and the effects of intermediate image generation.\ %
MindView evaluates generated views of observed scenes and tests whether target-view supervision aids spatial understanding.\ %

\textbf{Novel-View Synthesis and Viewpoint Control.}
View generation methods differ in target-view specification, from explicit camera parameters to language instructions.\ %
NeRF \citeblue{nerf2020}, ZeroNVS \citeblue{zeronvs2024}, EscherNet \citeblue{eschernet2024}, and LVSM \citeblue{lvsm2025} use supplied camera controls.\ %
CapFrame \citeblue{capframe2026} grounds language-specified viewpoints within a reconstructed 3D Gaussian scene, while Object-Uni \citeblue{objectuni2026} uses explicit object poses for controllable generation.\ %
Astra \citeblue{astra2026} generates language-guided imagined views for spatial question answering and also evaluates them against real targets.\ %
Existing methods either rely on explicit geometry or consider constrained language-specified viewpoints, leaving viewpoint inference and scene-consistent rendering from partial multi-view evidence underexplored.\ %

\textbf{Multi-Reference Image Generation and Editing.}
Multi-reference generation and editing use multiple input images to preserve and combine visual content.\ %
IP-Adapter \citeblue{ipadapter2023} and PhotoMaker \citeblue{photomaker2023} provide image- or identity-based conditioning, while MultiRef \citeblue{multiref2025} explicitly supports multiple visual references.\ %
MICE-Bench \citeblue{micebench2026} evaluates multi-reference creation and editing, while MultiBanana \citeblue{multibanana2026} tests generation from challenging combinations of visual references.\ %
Unlike these benchmarks, MindView treats inputs as partial observations of the same scene and evaluates language-guided target-view generation.\ %

%% file: sections/benchmark.tex
\section{MindView Benchmark}

In this section, we present MindView, a benchmark designed to evaluate
generative multi-view spatial intelligence.\ %
\Cref{sec:task-labels} formulates the target-view generation problem
and introduces seven tasks across two complementary families.\ %
\Cref{sec:construction-method} describes the data construction pipeline,
and \Cref{sec:evaluation-method} presents the evaluation protocol and metrics.\ %
 
\subsection{Task Formulation and Task Design}
\label{sec:task-labels}

MindView evaluates language-guided target-view generation from partial
observations of the same scene.\ %
Let $C_K=(I_1,\ldots,I_K)$ denote an ordered list of RGB observations and let
$q$ denote a spatial query.\ %
A generator $f$ predicts
\begin{equation}
    \hat I_{\mathrm{target}} = f(C_K, q).
    \label{eq:task}
\end{equation}
The primary benchmark setting uses $K=6$.\ %
At inference time, the generator receives only $C_K$ and $q$.\ %
The held-out real target image, camera poses, depth, and scene annotations are not
provided as generation inputs.\ %
When the query is defined relative to a starting viewpoint, $I_1$ serves as
the anchor.\ %
The generated image must preserve the observed scene while matching the
requested camera position and orientation.\ %

The seven tasks share this interface and differ in how the target viewpoint
is specified.\ %
\textbf{Geometry-Dominant Tasks} express it through egocentric camera motions
from an anchor.\ %
Translation (Trans.) changes camera position through relative displacements
while preserving heading.\ %
Rotation (Rot.) changes heading by a specified angle while keeping position
fixed.\ %
Composition (Comp.) combines translation and rotation in the order specified
by the query.\ %

\textbf{Semantics-Dominant Tasks} specify the target through referred objects,
spatial relations, and reference-frame cues.\ %
Relational Object Grounding (ROG) identifies the referred object through
relations and generates a view satisfying distance, centering, and visibility
constraints, without an anchor.\ %
Target-Facing Reorientation (TFR) requires locating the target object and
turning to face it while remaining at the anchor position.\ %
Object-Centric Repositioning (OCR) instead requires moving to a specified
side or angular offset around the object and facing it, with directions
defined relative to the anchor view.\ %
Multi-Cue Viewpoint Inference (MCVI) jointly infers position and heading
from observer-relative scene cues, without an anchor.\ %
Together, the two families test complementary forms of target-view
specification under a common generation interface.

\Cref{sec:task-examples} provides real examples and full generation
prompts for all seven tasks.\ %

\subsection{Data Construction}
\label{sec:construction-method}

\begin{figure*}[!t]
    \centering
    \includegraphics[width=\textwidth]{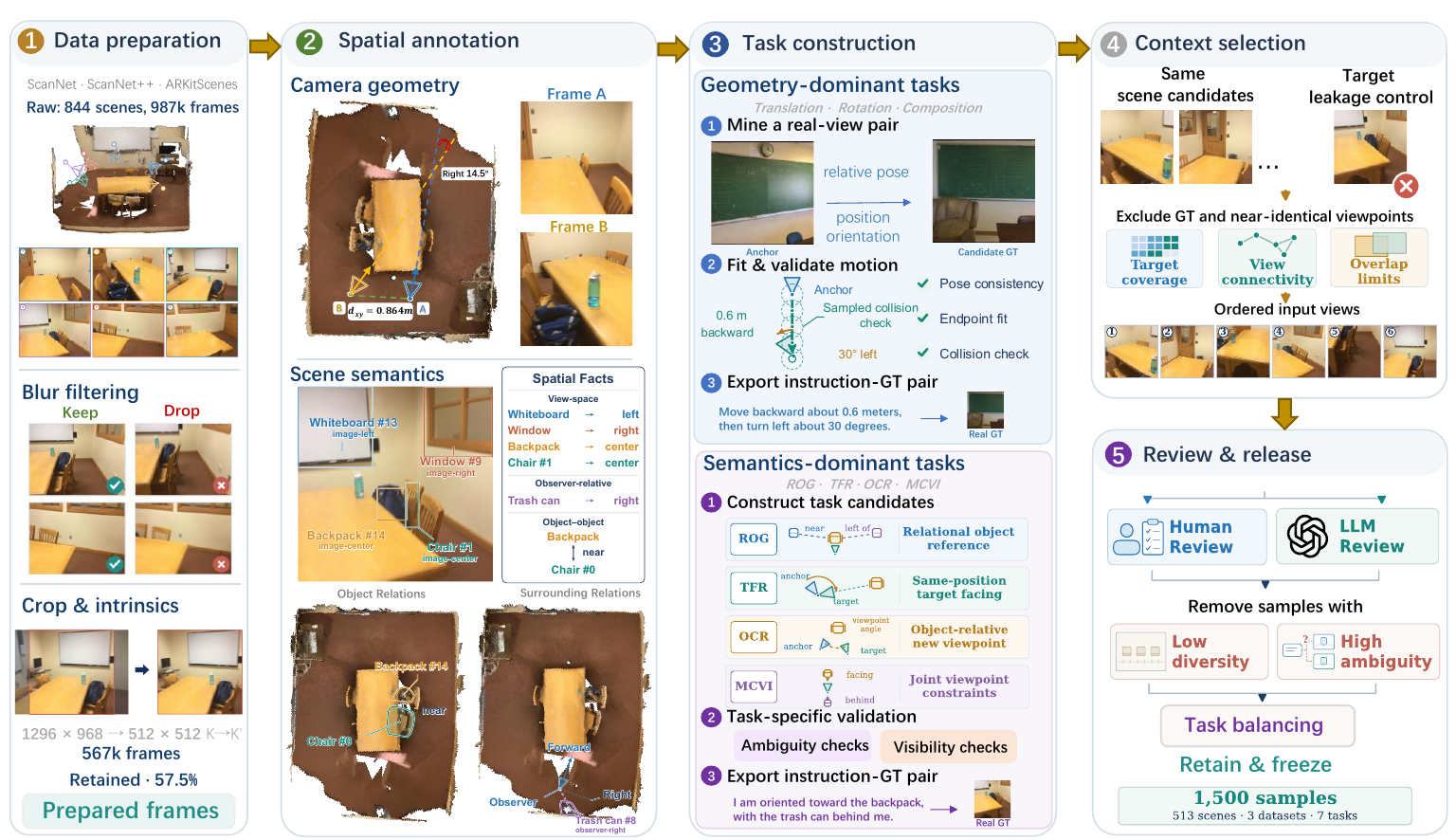}
    \caption{Five-stage offline construction pipeline for MindView.}
    \label{fig:data-construction-pipeline}
\end{figure*}

We construct MindView from ScanNet \citeblue{scannet2017},
ScanNet++ \citeblue{scannetpp2023}, and ARKitScenes
\citeblue{arkitscenes2021}.\ %
\Cref{fig:data-construction-pipeline} summarizes the five-stage
offline construction pipeline.\ %

\textbf{Data preparation and spatial annotation.}
We first apply blur filtering while preserving temporal coverage along each
recorded sequence.\ %
We then account for cropping and resizing in the camera model used for
geometric checks.\ %
Using camera poses, depth, scene geometry, and instance annotations, we derive
relative viewpoints, object relations, visibility, and object locations in
each image.\ %
These annotations form the spatial basis for constructing queries and targets.\ %

\textbf{Task construction.}
For Geometry-Dominant Tasks, relative poses between recorded views yield
motion instructions.\ %
Candidates are filtered by task-specific pose constraints, and endpoints
reconstructed from rounded instructions are checked against recorded targets.\ %
Where point clouds are available, Translation and Composition also use
discrete collision checks along instructed segments.\ %
For Semantics-Dominant Tasks, identities, relations, and reference-frame cues
define query-target candidates.\ %
Visibility and viewpoint checks limit ambiguity across recorded views.\ %
Each query uses a real captured target and, when required, an anchor.\ %
\Cref{sec:construction-details} details the task-specific criteria.\ %

\FloatBarrier
\textbf{Offline context selection.}
For each query-target pair, we select six unique same-scene views as a
connected path.\ %
The target is excluded, any required anchor comes first, and other candidates
pass temporal and pose-separation filters.\ %
Selection enforces pairwise-overlap limits and coverage of target-visible
content and the broader scene.\ %
MCVI also requires every object cue constraining observer location to appear
across the selected views.\ %

\begin{figure*}[!t]
    \centering
    \includegraphics[width=\textwidth]{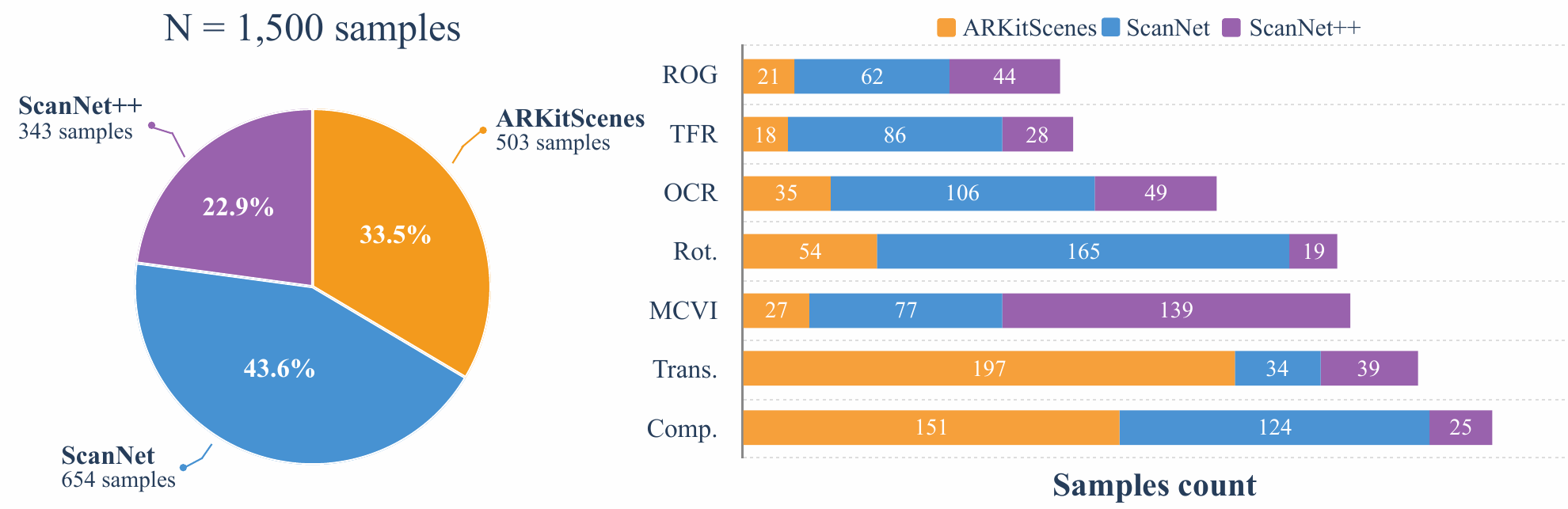}
    \caption{Distribution of the 1,500 MindView samples across source datasets and tasks.}
    \label{fig:dataset-statistics}
\end{figure*}

\textbf{Review and release.}
Candidate groups undergo human and LLM-assisted screening for residual
ambiguity and limited within-sample visual diversity.\ %
\Cref{sec:llm-review-prompt} provides the exact GPT-5.6 \citeblue{gpt562026} screening prompt.\ %
The frozen benchmark contains 1,500 samples from 513 scenes across the three
datasets (\Cref{fig:dataset-statistics}).\ %
\label{sec:training-data}
We also provide 20,000 training examples from 3,406 scenes, using the same
interface for all seven tasks.\ %
These examples use the source datasets' training splits and share no scenes
with the benchmark (\Cref{sec:training-resource-record}).\ %

\FloatBarrier

\subsection{Evaluation Protocol}
\label{sec:evaluation-method}

\textbf{Image-feature similarity.}
DINO-I probes overall visual-content fidelity through DINOv2-Large
feature correspondence \citeblue{dinov22023}.\ %
We use the cosine similarity between generated and real target features,
scaled by 100 without clipping.\ %

\textbf{VLM-based assessment.}
Qwen3-VL-30B-A3B-Instruct \citeblue{qwen3vl_judge_model} judges whether
scene content (\textbf{Semantic}) and visible viewpoint and layout
(\textbf{Spatial}) match the real target; \mbox{\textbf{Quality}} assesses
internal coherence and rendering defects in the generation alone.\ %
The judge receives no query or context; integer 0--10 ratings are scaled
to 0--100.\ %
See \Cref{sec:judge-protocol,sec:evaluator-prompts}
for inputs and prompts.\ %

\textbf{Geometric consistency.}
We introduce a VGGT-based protocol \citeblue{vggt2025} for \mbox{MindView}
to assess target-viewpoint fidelity through estimated camera position
and orientation.\ %
Joint inference on fixed references and generated ($\mathrm{gen}$) and
real ($\mathrm{real}$) target images yields rotations $\hat R$ and camera
centers $\hat{\mathbf c}$ in a shared frame.\ %
For sample $i$, the pose errors are
\begin{equation}
\bigl(e_{\mathrm{rot},i},e_{\mathrm{trans},i}\bigr)
=\left(
d_{\mathrm{SO}(3)}\!\left(\hat R_{\mathrm{gen}},\hat R_{\mathrm{real}}\right),\;
a_i\left\|\hat{\mathbf c}_{\mathrm{gen}}-\hat{\mathbf c}_{\mathrm{real}}\right\|_2
\right).
\label{eq:target-view-pose-errors}
\end{equation}
Here, $d_{\mathrm{SO}(3)}$ is angular distance in degrees, and $a_i>0$
converts VGGT lengths to meters using real-image sensor depths, without
ground-truth poses.\ %
Per-sample RScore and TScore decrease linearly from 100 at zero error to
0 at $60^\circ$ and $2\,\mathrm{m}$, respectively, and remain zero beyond
these thresholds.\ %
\Cref{sec:vggt-metric-protocol} details the reference inputs,
calibration, and scoring.\ %

All scores are higher-is-better.\ %
We average each component within tasks, then equally across the seven
tasks; Overall is the unweighted mean of the six task-macro scores.\ %

\textbf{Evaluator reliability.}
We assess evaluator reliability through comparisons with human ratings
and reconstruction stability tests.\ %
Three multimodal experts independently rate outputs from four generators.\ %
Aggregated human and VLM scores yield identical model rankings for
Semantic, Spatial, and Quality.\ %
For VGGT, we vary the generated image while keeping all real views fixed.\ %
Across samples, the median changes in the reconstructed target pose are
$0.71^\circ$ in relative rotation and 2.75\,cm in reference-aligned position.\ %
Together, these experiments support the reliability of our VLM-based and
geometric metrics for comparing generators (\Cref{sec:evaluator-reliability}).\ %

%% file: sections/experiments.tex
\section{Experiments}
\label{sec:experiments}

We first evaluate current generators on MindView and analyze their
task-level performance.\ %
We then examine whether additional observations improve target-view
generation.\ %
Motivated by these findings, we investigate task-specific adaptation through
supervised fine-tuning and training-free, query-guided evidence selection
through MindCue.\ %

\subsection{Experimental Setup}
\label{sec:experimental-setup}

\textbf{Evaluation set and models.}
All MindView generation comparisons use the frozen release of 1,500 aligned scene--query--target samples.\ %
The standard benchmark input consists of six context images and a language query per sample.\ %
All MindView generation results follow the evaluation protocol in \Cref{sec:evaluation-method}.\ %
We compare eleven off-the-shelf generators: seven open-source models
(BAGEL \citeblue{exp_bagel2025}, FLUX.2 \citeblue{exp_flux2_klein2026},
Qwen-Image-Edit-2511 \citeblue{exp_qwen_image2025,exp_qwen_image_edit2511},
Emu3.5 \citeblue{exp_emu35_2025}, DreamOmni2 \citeblue{exp_dreamomni2_2025},
SenseNova-U1 \citeblue{exp_sensenova_u1_2026}, and
SenseNova-U1.5 \citeblue{exp_sensenova_u15_2026})
and four closed-source models (Nano Banana \citeblue{exp_nanobanana2025},
Nano Banana Pro \citeblue{exp_nanobananapro2025},
GPT-Image-1.5 \citeblue{exp_gpt_image15_2025}, and
GPT-Image-2 \citeblue{exp_gpt_image2_2026}).\ %
We additionally evaluate BAGEL-SFT, obtained by supervised fine-tuning
of BAGEL on our curated training dataset for 10,000 steps.\ %

\subsection{Benchmark results}
\label{sec:baseline-results}

% Declare result floats in their owning subsection; all use top-only placement.
\input{tables/base_generators_c6_overall_metrics}
\begin{figure*}[!t]
    \centering
    \includegraphics[width=\textwidth]{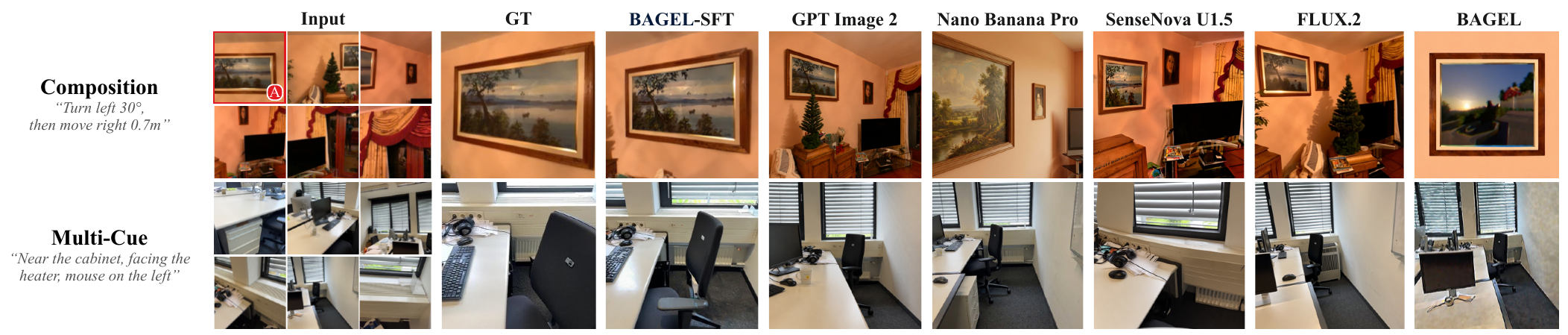}
    \caption{Qualitative comparison on Composition and Multi-Cue Viewpoint Inference.}
    \label{fig:baseline-qualitative}
\end{figure*}

\textbf{Quantitative results.}
Among the eleven off-the-shelf generators, GPT-Image-2 achieves the highest
Overall score (57.9), while SenseNova-U1.5 leads the open-source group
(51.3; \Cref{tab:final-c6-overall}).\ %
GPT-Image-2 also leads on all seven tasks, but task-level weaknesses recur
across models: Composition is the lowest-scoring task for eight generators,
and Multi-Cue Viewpoint Inference is lowest for the remaining three
(\Cref{tab:c6-results}).\ %
These results highlight recurring difficulties with composed camera motions
and viewpoints constrained by multiple scene cues.\ %

Content fidelity and geometric consistency also yield different model rankings.\ %
Nano Banana Pro scores higher than SenseNova-U1.5 on DINO-I and Semantic,
yet lower on RScore (38.8 versus 41.7) and TScore (39.0 versus 43.2).\ %
This discrepancy underscores the value of evaluating target-view geometry
alongside content fidelity when assessing language-guided view generation.\ %

\textbf{Qualitative results.}
Consistent with the quantitative findings,
\Cref{fig:baseline-qualitative} illustrates viewpoint and content
errors across models on two challenging tasks, Composition and Multi-Cue
Viewpoint Inference.\ %
In Composition, Nano Banana Pro emphasizes the painting but differs from
the target in both viewing angle and the painting's content.\ %
In Multi-Cue Viewpoint Inference, GPT-Image-2 and FLUX.2 include the desk,
chair, and heater, but depict the chair as smaller and farther back in the
image than in the target view.\ %
Appendix \Cref{fig:appendix-baseline-qualitative} provides comparisons
across all seven tasks.\ %

\input{tables/base_generators_c6_task_composite_scores}

\FloatBarrier
\subsection{Context Scaling and Evidence Utilization}
\label{sec:context-results}

\input{tables/context_budget_composite_scores}

We examine context expansion on the same 1,500 scene--query--target
samples using nested \mbox{three-,} six-, and ten-view settings, with $C_6$
as the standard input.\ %
$C_3$ is an ordered subsequence of $C_6$, which is an ordered
subsequence of $C_{10}$.\ %
Target images are excluded from generator inputs, and VGGT uses the
same ten evaluation references across settings.\ %
Computed only for offline analysis, target coverage measures the fraction
of content visible in the real target that is also observed in at least
one context view.\ %
Its sample mean rises from 57.6\% at $C_3$ to 88.4\% at $C_6$ and
92.0\% at $C_{10}$.\ %
Despite this increase, all eleven off-the-shelf generators and BAGEL-SFT
obtain lower Overall at $C_{10}$ than at $C_6$
(\Cref{fig:context-budget}).\ %
Greater coverage therefore does not ensure better generation under this
protocol.\ %
This observation motivates MindCue, which selects query-relevant evidence
from the four $C_{10}$ additions while retaining the complete $C_6$ core
(\Cref{sec:mindcue}).\ %
Full component scores appear in Appendix
\Cref{tab:appendix-context-metrics}.\ %

\FloatBarrier
\subsection{Supervised Adaptation and Transfer to Spatial Understanding}
\label{sec:sft-results}

\textbf{Supervised adaptation.}
To examine whether task-specific supervision improves target-view
generation, we fine-tune BAGEL \citeblue{exp_bagel2025} on our curated
training dataset for 10,000 steps.\ %
Adaptation raises Overall from 40.0 to 64.4, a gain of 24.3 points,
as shown in \Cref{tab:final-c6-overall}, and all six metrics improve
on every task (Appendix
\Crefrange{tab:baseline-task-components-1}{tab:baseline-task-components-4}).\ %
The adapted model's task-macro advantage over GPT-Image-2 is larger
on geometry-dominant than semantics-dominant tasks
(10.8 versus 3.3 points; \Cref{tab:c6-results}), possibly because explicit motion
supervision provides more regular patterns to learn, whereas
semantics-dominant tasks demand more complex semantic
interpretation and spatial reasoning.\ %
In the Composition example in \Cref{fig:baseline-qualitative},
BAGEL-SFT more closely matches the GT's painting-dominated close-up
than GPT-Image-2, which instead produces a broad room view.\ %
% Quality nevertheless remains below GPT-Image-2 (61.8 versus 84.6).\ %
% Since this metric evaluates intrinsic image quality rather than
% task-specific viewpoint correctness, the gap may partly reflect
% differences in the underlying models' rendering capabilities.\ %
Training details are provided in \Cref{sec:sft-details}.\ %

\input{tables/sft_spatial_understanding_transfer}

% Place the MindCue overview directly below the transfer table.
\begin{figure*}[!t]
    \centering
    \setlength{\abovecaptionskip}{6pt}
    \includegraphics[width=\textwidth,trim=0 4bp 0 0,clip]{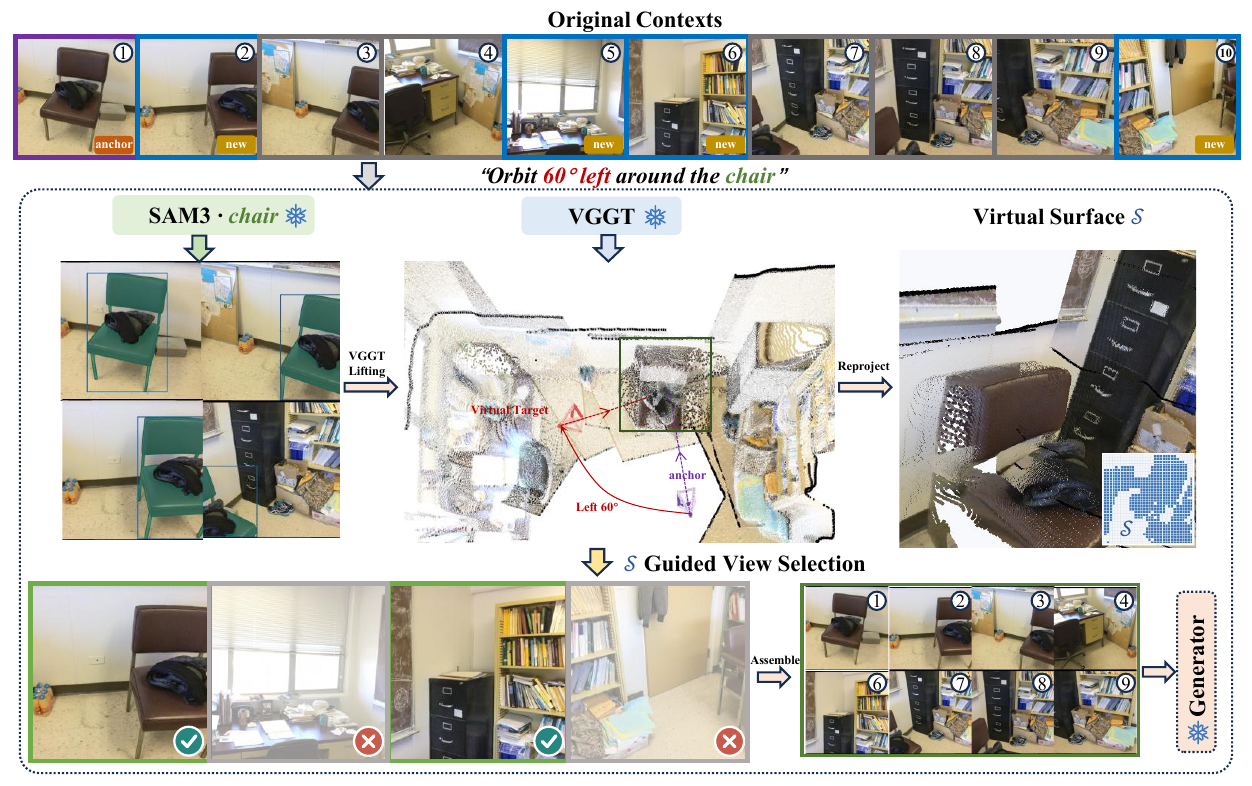}
    \caption{Overview of the MindCue framework.}
    \label{fig:mindcue-pipeline}
\end{figure*}

\textbf{Transfer to spatial understanding.}
\label{sec:spatial-transfer}
We next examine whether learning to generate target views also benefits
spatial understanding.\ %
To this end, we evaluate BAGEL and BAGEL-SFT on SAT-v2 \citeblue{exp_sat2025},
MMSI-Bench \citeblue{exp_mmsi2026}, and MindCube-Tiny
\citeblue{exp_mindcube2025}, strictly following their respective
input-construction and answer-scoring protocols
(\Cref{sec:transfer-details}).\ %
Our fine-tuning stage uses no annotations from these benchmarks.\ %
All results are obtained in our own evaluations.\ %
Our reproduced BAGEL accuracies on MMSI-Bench and MindCube-Tiny match
those reported by SenseNova-SI \citeblue{exp_sensenova_si2026}
at the reported precision.\ %

As shown in \Cref{tab:spatial-understanding}, BAGEL-SFT obtains
higher scores in all three comparisons.\ %
On SAT-v2, micro accuracy increases from 58.0\% to 62.7\%, while
both-order accuracy rises from 27.3\% to 41.3\%, indicating that more
questions are answered correctly under both evaluated answer orders.\ %
Its scores on MMSI-Bench and MindCube-Tiny also exceed the reproduced BAGEL
baselines by 1.2 and 1.8 percentage points, respectively.\ %
Together, these results provide preliminary evidence that learning to
generate spatially grounded views may also benefit spatial understanding,
even without direct supervision from the evaluated benchmarks.\ %

\FloatBarrier
\subsection{MindCue: A Training-Free Spatial Evidence Selector}
\label{sec:mindcue}

Motivated by the context-scaling results, \textbf{MindCue} uses a target-view
hypothesis to select additional spatial evidence without updating model weights
(\Cref{fig:mindcue-pipeline}).\ %

\textbf{Method.}
For sample $i$, let $C_i^{\mathrm{pool}}$ be the ordered observations,
$q_i$ the query, and $\mathcal F_i,\mathcal D_i$ the core and additional-view
index sets.\ %
Frozen VGGT \citeblue{vggt2025} provides confidence-filtered scene points
$\hat{\mathcal P}_i$; for semantics-dominant tasks, frozen SAM~3
\citeblue{sam32026} segments queried entities.\ %
A task-specific solver uses this evidence, the query, and core/reference-view
metadata to hypothesize a target camera $\hat{\mathcal C}^{\star}_i$
(pose and intrinsics).\ %
Its virtual surface is
\begin{equation}
\mathcal S_i
  =\operatorname{Occ}\!\left[
    \operatorname{ZBuf}\!\left(
    \Pi_{\hat{\mathcal C}^{\star}_i}\!\left(\hat{\mathcal P}_i\right)
    \right)\right].
\label{eq:mindcue-target-surface}
\end{equation}
Here $\Pi$ projects points, $\operatorname{ZBuf}$ retains the nearest valid depth
per grid cell, and $\operatorname{Occ}$ returns the occupied cells.\ %

For nonempty $\mathcal S_i$, let $R_i(v)$ measure additional view $v$'s reliable coverage,
capping credit for geometry supported only by that view, and let $\tau$
be the coverage threshold.\ %
$X_i\subseteq\mathcal D_i$ contains at most one below-threshold candidate
passing connectivity and nonredundancy tests; without an explicit anchor,
it is empty.\ %
The retained additional-view indices are
\begin{equation}
A_i=
\begin{cases}
\varnothing, & \mathcal S_i=\varnothing,\\
\left\{v\in\mathcal D_i:R_i(v)\geq\tau\right\}\cup X_i,
  & \mathcal S_i\neq\varnothing.
\end{cases}
\label{eq:mindcue-selection}
\end{equation}
The generator receives the original query and pool views indexed by
$\mathcal F_i\cup A_i$, preserving their order and the complete core.\ %
Selection uses no held-out targets, ground-truth geometry, generated outputs,
or evaluation scores.\ %
\Cref{sec:selection-details} specifies the support and selection rules;
\Cref{tab:mindcue-camera-solvers} details the task-specific cameras.\ %

\textbf{Evaluation.}
We compare vanilla and fine-tuned BAGEL on the same 1,500 samples with fixed
checkpoints and no-thinking inference.\ %
Fixed $C_6$ and $C_{10}$ supply the core and full pool, respectively.\ %
Both controls preserve the core: RandDrop randomly selects the same number of
additions as MindCue per sample, while CounterSel retains rejected additions
and is not count-matched.\ %
\Cref{tab:training_free_effectiveness} reports Overall and view counts;
Appendix \Cref{tab:selection-component-metrics} gives all six components.\ %

\input{tables/training_free_context_selection_results}

\textbf{Results.}
MindCue exceeds both fixed-context baselines on BAGEL and BAGEL-SFT,
reversing the expanded-context decline and improving on the standard input.\ %
Its gains over count-matched RandDrop are 1.6 and 1.1 points for vanilla
and fine-tuned BAGEL, respectively, supporting selection beyond count reduction.\ %
The improvement over RandDrop extends to five of the six component scores
for BAGEL and all six for BAGEL-SFT.\ %
CounterSel scores lowest among all five policies for both models, consistent
with the value of target-guided retention.\ %

% BEGIN MEASURED MINDCUE RUNTIME
\textbf{Runtime.}
On a single A100 80GB, MindCue adds 1.44~s on average, or 3.4\% of BAGEL
inference time with identical selected views and input order.\ %
The task-averaged costs of VGGT reconstruction and SAM~3 segmentation with
mask lifting are 0.57~s and 0.82~s, respectively.\ %
SAM~3 is used only for semantics-dominant tasks.\ %
The remaining selection steps add 0.05~s.\ %
We sum separately timed stages with preloaded models, excluding preprocessing
(\Cref{sec:mindcue-runtime}).\ %
Thus, query-guided selection adds only modest computational overhead
relative to generation.\ %
% END MEASURED MINDCUE RUNTIME

%% file: tables/base_generators_c6_overall_metrics.tex
\begin{table*}[!t]
\centering
\caption{\textbf{Quantitative comparison on MindView.}
Task-macro scores over 1,500 samples.
\protect\raisebox{0.10em}{\protect\textcolor{MindDirectPale}{\protect\rule{0.60em}{0.60em}}} and
\protect\raisebox{0.10em}{\protect\textcolor{MindTealPale}{\protect\rule{0.60em}{0.60em}}} mark best and
second-best across displayed models, respectively; \textsuperscript{\ensuremath{\dagger}} denotes tiled context views.}
\label{tab:final-c6-overall}
\label{tab:appendix-c6-metrics}
\begingroup
\fontsize{8}{9}\selectfont
\setlength{\tabcolsep}{4.0pt}
\setlength{\arrayrulewidth}{0.4pt}
\renewcommand{\arraystretch}{1.00}
\newcommand{\scorebadge}[2]{\begingroup\setlength{\fboxsep}{0.65pt}\colorbox{#1}{#2}\endgroup}
\newcommand{\bestscore}[1]{\scorebadge{MindDirectPale}{\textbf{#1}}}
\newcommand{\secondscore}[1]{\scorebadge{MindTealPale}{#1}}

\begin{tabular*}{\textwidth}{@{\extracolsep{\fill}}c|c|ccc|cc|c@{}}
\toprule
\multicolumn{1}{c|}{\multirow{2}{*}{\textbf{Model}}}
& \multicolumn{1}{c|}{\multirow{2}{*}{\textbf{DINO-I}}}
& \multicolumn{3}{c|}{\textbf{VLM Judge}}
& \multicolumn{2}{c|}{\textbf{Geometry}}
& \multirow{2}{*}{\textbf{Overall}} \\
\cmidrule(lr){3-5}\cmidrule(lr){6-7}
& & \textbf{Semantic} & \textbf{Spatial}
& \multicolumn{1}{c|}{\textbf{Quality}}
& \textbf{RScore} & \multicolumn{1}{c|}{\textbf{TScore}} & \\
\midrule
\rowcolor{MindGroundedPale}
\multicolumn{8}{c}{\strut\bfseries\itshape Open-Source Models} \\
BAGEL & 48.3 & 44.5 & 33.6 & 47.1 & 34.7 & 31.9 & 40.0 \\
FLUX.2 & 52.2 & 49.5 & 39.4 & 80.4 & 35.5 & 38.6 & 49.3 \\
Qwen-Image-Edit-2511\textsuperscript{\ensuremath{\dagger}} & 52.4 & 49.2 & 38.6 & 69.5 & 37.2 & 40.3 & 47.9 \\
Emu3.5\textsuperscript{\ensuremath{\dagger}} & 44.9 & 41.0 & 31.2 & 67.2 & 36.0 & \secondscore{49.9} & 45.0 \\
DreamOmni2 & 41.7 & 36.0 & 28.8 & 65.5 & 25.2 & 42.7 & 40.0 \\
SenseNova-U1 & 53.6 & 48.2 & 38.3 & 63.5 & 25.6 & 33.2 & 43.7 \\
SenseNova-U1.5 & 58.0 & 52.8 & 42.6 & 69.7 & 41.7 & 43.2 & 51.3 \\

\rowcolor{MindAmberPale}
\multicolumn{8}{c}{\strut\bfseries\itshape Closed-Source Models} \\
Nano Banana & 56.5 & 54.0 & 41.7 & 66.2 & 39.0 & 40.7 & 49.7 \\
Nano Banana Pro & 60.0 & 57.2 & 45.4 & 80.0 & 38.8 & 39.0 & 53.4 \\
GPT-Image-1.5 & 58.8 & 58.9 & 46.3 & \secondscore{82.2} & 39.9 & 33.5 & 53.3 \\
GPT-Image-2 & \secondscore{63.0} & \secondscore{61.5} & \secondscore{49.0} & \bestscore{84.6} & \secondscore{49.9} & 39.1 & \secondscore{57.9} \\

\rowcolor{MindAdaptPale}
\multicolumn{8}{c}{\strut\bfseries\itshape Task-Specific Adaptation} \\
BAGEL-SFT & \bestscore{68.0} & \bestscore{71.0} & \bestscore{57.1} & 61.8 & \bestscore{61.7} & \bestscore{66.5} & \bestscore{64.4} \\
\bottomrule
\end{tabular*}

\endgroup
\end{table*}

%% file: tables/base_generators_c6_task_composite_scores.tex
\begin{table*}[!t]
\centering
\caption{\textbf{Task-wise comparison on MindView.}
Per-task Overall scores across all displayed models.
Highlighting and \textsuperscript{\ensuremath{\dagger}} follow \Cref{tab:final-c6-overall};
abbreviations are defined in \Cref{sec:task-labels}.}
\label{tab:c6-results}
\label{tab:final-c6-taskwise}
\begingroup
\fontsize{8}{9}\selectfont
\setlength{\tabcolsep}{2.3pt}
\setlength{\arrayrulewidth}{0.4pt}
\renewcommand{\arraystretch}{1.00}
\newcommand{\scorebadge}[2]{\begingroup\setlength{\fboxsep}{0.65pt}\colorbox{#1}{#2}\endgroup}
\newcommand{\bestscore}[1]{\scorebadge{MindDirectPale}{\textbf{#1}}}
\newcommand{\secondscore}[1]{\scorebadge{MindTealPale}{#1}}

\begin{tabular*}{\textwidth}{@{\extracolsep{\fill}}c|ccc|cccc@{}}
\toprule
\multicolumn{1}{c|}{\multirow{2}{*}{\textbf{Model}}}
& \multicolumn{3}{c|}{\textbf{Geometry-Dominant Tasks}}
& \multicolumn{4}{c}{\textbf{Semantics-Dominant Tasks}} \\
\cmidrule(lr){2-4}\cmidrule(lr){5-8}
& \textbf{Trans.} & \textbf{Rot.} & \multicolumn{1}{c|}{\textbf{Comp.}}
& \textbf{ROG} & \textbf{TFR} & \textbf{OCR} & \textbf{MCVI} \\
\midrule
\rowcolor{MindGroundedPale}
\multicolumn{8}{c}{\strut\bfseries\itshape Open-Source Models} \\
BAGEL & 41.7 & 38.8 & 36.1 & 43.2 & 38.7 & 44.1 & 37.6 \\
FLUX.2 & 53.0 & 47.0 & 44.1 & 50.5 & 49.9 & 53.9 & 46.6 \\
Qwen-Image-Edit-2511\textsuperscript{\ensuremath{\dagger}} & 53.4 & 45.9 & 41.6 & 49.9 & 46.6 & 57.0 & 40.7 \\
Emu3.5\textsuperscript{\ensuremath{\dagger}} & 55.7 & 43.6 & 40.7 & 42.5 & 45.0 & 52.3 & 35.5 \\
DreamOmni2 & 54.4 & 40.9 & 35.8 & 33.7 & 35.0 & 47.9 & 32.2 \\
SenseNova-U1 & 44.2 & 40.1 & 37.9 & 49.4 & 46.8 & 44.6 & 43.3 \\
SenseNova-U1.5 & 53.6 & 46.8 & 43.2 & 56.7 & 53.4 & 54.7 & 50.9 \\

\rowcolor{MindAmberPale}
\multicolumn{8}{c}{\strut\bfseries\itshape Closed-Source Models} \\
Nano Banana & 49.3 & 48.1 & 45.8 & 53.4 & 55.7 & 48.3 & 47.3 \\
Nano Banana Pro & 54.5 & 53.7 & 47.3 & 56.3 & 58.6 & 54.7 & 48.7 \\
GPT-Image-1.5 & 52.1 & 47.9 & 46.3 & 59.2 & 59.4 & 54.3 & 53.7 \\
GPT-Image-2 & \secondscore{57.0} & \secondscore{54.2} & \secondscore{48.2} & \secondscore{59.4} & \bestscore{66.8} & \secondscore{62.4} & \bestscore{57.0} \\

\rowcolor{MindAdaptPale}
\multicolumn{8}{c}{\strut\bfseries\itshape Task-Specific Adaptation} \\
BAGEL-SFT & \bestscore{67.1} & \bestscore{68.2} & \bestscore{56.3} & \bestscore{62.7} & \secondscore{66.5} & \bestscore{73.4} & \secondscore{56.2} \\
\bottomrule
\end{tabular*}

\endgroup
\end{table*}

%% file: tables/context_budget_composite_scores.tex
\begin{figure*}[!t]
    \centering
    \setlength{\abovecaptionskip}{6pt}
    \includegraphics[width=\textwidth,trim=0 16bp 0 0,clip]{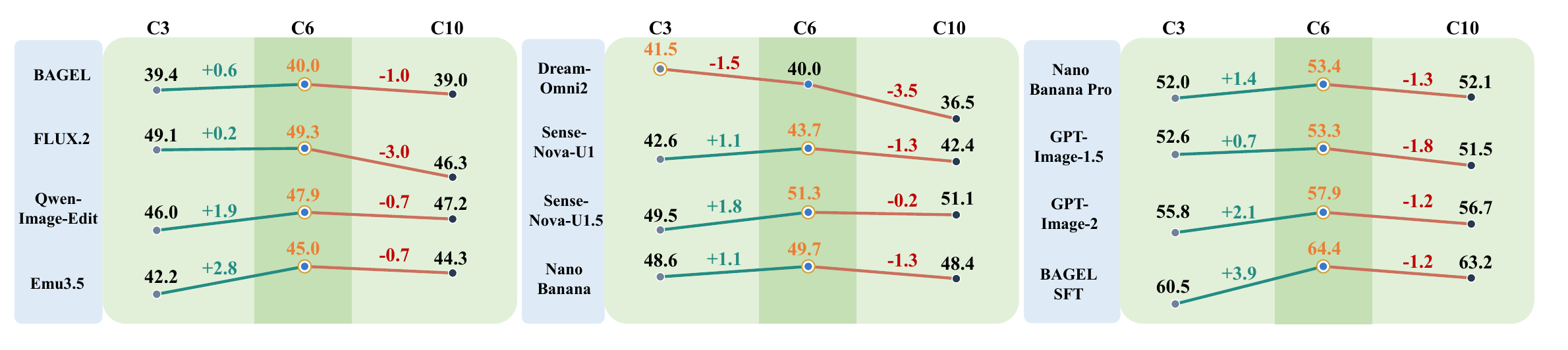}
    \caption{Overall score trajectories across nested context settings.}
    \label{fig:context-budget}
\end{figure*}

%% file: tables/sft_spatial_understanding_transfer.tex
\begin{table}[!t]
\centering
\setlength{\abovecaptionskip}{4pt}
\caption{\textbf{Spatial-understanding transfer.}
We report accuracy (\%) and gains in percentage points.}
\label{tab:spatial-understanding}
\begingroup
\footnotesize
\setlength{\tabcolsep}{3.5pt}
\setlength{\arrayrulewidth}{0.4pt}
\renewcommand{\arraystretch}{1.03}
\newcommand{\transferdiff}[1]{\textcolor{MindDirect}{\textbf{#1}}}
\newsavebox{\spatialunderstandingtablebox}

\begin{lrbox}{\spatialunderstandingtablebox}
\begin{tabular}{@{}cccccc@{}}
\toprule
\multirow{2}{*}{\textbf{Benchmark}}
& \multirow{2}{*}{\textbf{Protocol}}
& \multirow{2}{*}{\textbf{Eval. Size}}
& \multicolumn{2}{c}{\textbf{Accuracy (\%)}}
& \multirow{2}{*}{\makecell{\textbf{Difference}\\\textbf{(pp)}}} \\
\cmidrule(lr){4-5}
& & & \textbf{BAGEL} & \textbf{BAGEL-SFT} & \\
\midrule
\multirow{2}{*}{SAT-v2}
& Micro accuracy & 300 & 58.0 & 62.7 & \transferdiff{+4.7} \\
& Both-order accuracy & 150 & 27.3 & 41.3 & \transferdiff{+14.0} \\
\addlinespace[2pt]
MMSI-Bench
& Vanilla accuracy & 1,000 & 31.0 & 32.2 & \transferdiff{+1.2} \\
\addlinespace[2pt]
MindCube-Tiny
& Raw-QA accuracy & 1,050 & 34.7 & 36.5 & \transferdiff{+1.8} \\
\bottomrule
\end{tabular}
\end{lrbox}
\usebox{\spatialunderstandingtablebox}

\endgroup
\end{table}

%% file: tables/training_free_context_selection_results.tex
\begin{table}[!t]
\centering
\caption{\textbf{Task-macro Overall under context selection.}
Parentheses give mean view counts;
\protect\mbox{\protect\raisebox{0.10em}{\protect\textcolor{MindDirectPale}{\protect\rule{0.60em}{0.60em}}} and
\protect\raisebox{0.10em}{\protect\textcolor{MindTealPale}{\protect\rule{0.60em}{0.60em}}}} mark the best and
second-best policies per model, respectively.
$\Delta C_k$ is MindCue's score-point gain over Fixed $C_k$, computed before rounding.}
\label{tab:training_free_effectiveness}
\begingroup
\footnotesize
\setlength{\tabcolsep}{5.0pt}
\setlength{\arrayrulewidth}{0.4pt}
\renewcommand{\arraystretch}{1.15}
\newcommand{\scorebadge}[2]{\begingroup\setlength{\fboxsep}{1.1pt}\colorbox{#1}{#2}\endgroup}
\newcommand{\bestscore}[1]{\scorebadge{MindDirectPale}{\textbf{#1}}}
\newcommand{\secondscore}[1]{\scorebadge{MindTealPale}{#1}}
\newcommand{\mindcuegain}[1]{\textcolor{MindTeal!75!black}{#1}}

\begin{tabular}{@{}c|ccccc|cc@{}}
\toprule
\textbf{Model}
& \makecell{\textbf{Fixed $C_6$}\\(6.0)}
& \makecell{\textbf{Fixed $C_{10}$}\\(10.0)}
& \makecell{\textbf{CounterSel}\\(8.6)}
& \makecell{\textbf{RandDrop}\\(7.4)}
& \makecell{\textbf{MindCue}\\(7.4)}
& \textbf{$\Delta C_6$} & \textbf{$\Delta C_{10}$} \\
\midrule
BAGEL & \secondscore{40.0} & 39.0 & 37.1 & 39.6 & \bestscore{41.1} & \mindcuegain{+1.1} & \mindcuegain{+2.1} \\
BAGEL-SFT & \secondscore{64.4} & 63.2 & 62.6 & 64.1 & \bestscore{65.1} & \mindcuegain{+0.8} & \mindcuegain{+1.9} \\
\bottomrule
\end{tabular}
\endgroup
\end{table}

%% file: sections/conclusion.tex
\section{Conclusion}

We introduce MindView, a benchmark for generative multi-view spatial
intelligence through language-guided target-view generation from partial
multi-view observations.\ %
Evaluation of eleven generators reveals challenges in content fidelity
and viewpoint consistency, while expanding from six to ten views lowers
Overall for all eleven.\ %
Fine-tuning BAGEL on our training set improves generation across all seven
tasks and provides preliminary evidence of transfer to spatial understanding.\ %
Complementing task-specific adaptation, MindCue uses a target-view
hypothesis for training-free evidence selection and improves both
BAGEL variants over standard and expanded inputs and count-matched
random selection.\ %
These findings highlight task learning and query-guided
evidence selection as complementary directions for advancing
generative multi-view spatial intelligence.\ %

%% file: sections/statements.tex
\subsection*{AI use statement}
We used AI tools to polish the language and improve the presentation of
the manuscript.\ %
GPT-5.6 screened and filtered candidate benchmark samples for target
ambiguity and insufficient context-view diversity before expert review
(\Cref{sec:context-review-release}).\ %
Our use of Qwen3-VL for model-based evaluation is described in
\Cref{sec:judge-protocol}.\ %
The authors take responsibility for the final manuscript.\ %

\subsection*{Reproducibility statement}
We will release the data, code, and detailed documentation needed for
faithful reproduction of our main experimental findings.\ %
\Cref{sec:construction-details} details benchmark construction and
review, including the exact screening prompt.\ %
\Cref{sec:metric-details} specifies the evaluation metrics,
aggregation rules, and evaluator prompts.\ %
Training and inference settings, spatial-understanding evaluation protocols,
and MindCue implementation are provided in
\Cref{sec:sft-details,sec:transfer-details,sec:selection-details}, respectively.\ %

%% file: sections/appendix.tex
% Top-align float-only pages and retain normal spacing between their floats.
\makeatletter
\setlength{\@fptop}{0pt}
\setlength{\@fpsep}{\floatsep}
\makeatother

\section{Benchmark Construction Details}
\label{sec:construction-details}

This section expands the five offline stages in
\Cref{sec:construction-method}.
Task-specific constructors may interleave candidate generation, query
construction, and validation.\ %
The generator receives none of these construction signals and sees only
selected RGB views and the query.\ %

\subsection{Data Preparation and Spatial Annotation}

\textbf{Data preparation.}
We apply the same blur-filtering procedure to ScanNet, ScanNet++, and
\mbox{ARKitScenes}.\ %
We convert each frame to grayscale and resize its long side to 320 pixels.\ %
Its \mbox{Tenengrad} sharpness score is the mean squared gradient magnitude
computed with $3\times3$ Sobel filters \citeblue{santos1997autofocus}.\ %
Within each sequence, we retain frames at or above the median
sharpness score and add the highest-scoring frame from each non-overlapping
five-frame window.\ %
This union favors image clarity while preserving temporal coverage.\ %
For geometric checks, we use a centered square region of each retained
image and adjust its camera intrinsics for resizing and cropping.\ %
Projection and visibility checks use these adjusted intrinsics and the
corresponding image coordinates.\ %

\textbf{Spatial annotation.}
Recorded camera poses provide relative translations and heading changes.\ %
Horizontal displacements are expressed in a ground-plane frame aligned
with the relevant camera heading.\ %
Instance labels and 3D geometry provide object identities, locations,
extents, and spatial relationships, including proximity and vertical
arrangement.\ %
We also derive object directions and distances relative to the observer.\ %
For each view, we project instance points using its pose and adjusted
intrinsics, then compare their depths with recorded depth to estimate
visibility within the effective image crop.\ %
The visible points support image-space locations and view-dependent
object relationships.\ %
These annotations support query construction and target-view selection.\ %

\subsection{Task-Specific Construction}

Using these annotations, we construct each task around the information that
specifies its target viewpoint.\ %
Geometry-Dominant Tasks convert the relative pose between recorded views into
a motion query, whereas Semantics-Dominant Tasks use objects and spatial cues
to identify valid target views.\ %
Every query has a real recorded target and uses an anchor only when it is
defined relative to a starting view.\ %
Task-specific checks enforce geometric fidelity and, for semantic queries,
reduce ambiguity across recorded views.\ %

\textbf{Geometry-Dominant Tasks.}
Translation encodes one or two ordered horizontal steps in the anchor camera
frame without a commanded turn.\ %
Rotation changes the horizontal heading with only small camera-center drift.\ %
Composition supports either motion order.\ %
A displacement uses the anchor frame before a turn and the rotated camera
frame after it.\ %
We retain pairs that satisfy task-specific position and heading constraints,
together with height, pitch, and roll gates.\ %
Movement distances are rounded and turn angles are quantized for the query.\ %
We execute the discrete instruction geometrically and require the resulting
camera center to remain close to the recorded target position.\ %
For Translation and Composition, available point clouds support collision
checks with vertical body cylinders sampled along each translated segment.\ %
We reject a candidate if any sampled cylinder reaches the occupancy threshold,
using these discrete checks to assess the traversability of the instructed path.\ %

\textbf{Relational Object Grounding.}
Without an anchor, the query uses one to three clear relational facts to
identify an object.\ %
Its target is a real view in which that object is near the image center and all
referenced objects are sufficiently visible.\ %
We remove views with a clear same-category distractor and reject a description
that also matches another valid target candidate.\ %
Views supporting the same description are grouped when their camera poses and
projected object layouts are equivalent.\ %
We select among these clusters by the horizontal clearance from the camera to
the target object's footprint and reject cases with a near-tied alternative.\ %
ScanNet++ and \mbox{ARKitScenes} allow a single-relation fallback only if no
multi-relation candidate exists and the scene records match the description to
exactly one target instance and one compatible view cluster.\ %
This fallback occurs in 44 of 127 released ROG queries.\ %
The final instruction also specifies target distance and centering and requires
all named objects to remain visible.\ %

\textbf{Target-Facing Reorientation.}
The task asks for an approximately in-place turn from Image~1 toward a uniquely
described object.\ %
The object phrase must identify one scene instance, using either its category
alone or that category with at most one nearby reference.\ %
We retain a real target view only when its camera faces the object and the
object is sufficiently visible, fully inside the effective crop, and near the
image center.\ %
A target is rejected when a similarly suitable view from approximately the
same position points in a substantially different direction.\ %
We then select an anchor with a nearby camera center, similar height, a
different heading, and consistent roll.\ %
To ensure that the turn reveals the requested object, we exclude anchors in
which it passes a lenient depth-aware visibility test.\ %

\textbf{Object-Centric Repositioning.}
The task begins with an anchor that clearly shows the reference object near the
image center, without another salient instance of the same category.\ %
We then select a target at the requested left or right angular offset, or on
the side of the object opposite the anchor, with left and right defined in the
anchor camera frame.\ %
The target must face and center the object from a similar distance while
remaining consistent with the anchor in height, pitch, and roll.\ %
We discard queries when comparably suitable candidates imply distinct target
poses.\ %
These checks validate the destination view, not a collision-free path around
the object.\ %

\textbf{Multi-Cue Viewpoint Inference.}
Without an anchor, the query specifies the desired viewpoint through the
observer's position relative to nearby objects and the image locations of
visible objects.\ %
We match the complete cue set against every valid annotated frame in the scene
and retain it only when all matches form one compact cluster in horizontal
position and heading.\ %
A representative real frame from this cluster serves as the target.\ %
Within each scene, we remove repeated cue signatures and near-duplicate target
poses.\ %
Because the search covers only recorded frames, it tests uniqueness among
annotated observations rather than over continuous camera space.\ %

Together, these procedures yield validated query-target pairs whose contexts
are selected in the next stage.\ %

% The screening prompt begins on the next page.  Keep the resulting short page
% from stretching the subsection spacing, then restore the template default.
\raggedbottom
\subsection{Context Selection, Review, and Release}
\label{sec:context-review-release}

\textbf{Offline context selection.}
The selector assembles six unique same-scene frames and orders them as a
connected path, with any required anchor first.\ %
The exact target is excluded.\ %
Non-anchor candidates must satisfy temporal separation and are filtered when
both horizontal position and viewing direction are too close to the target.\ %
Pairwise-overlap limits and coverage of target-visible content and the broader
scene are enforced.\ %
For MCVI, we additionally exclude views identified as equivalent answers
during query validation.\ %

\textbf{Human and LLM-assisted review.}
We first screened 2,227 candidate groups with GPT-5.6 for residual target
ambiguity and limited context-view diversity.\ %
The model received the task type, query, six labeled context images, held-out
target, and anchor indicator, and returned a binary decision.\ %
The prompt restricted target use to consistency checks, not ambiguity
resolution.\ %
This stage retained 2,062 groups and rejected 165.\ %
The exact LLM screening prompt appears below.\ %

Three experts in multimodal research then independently reviewed all 2,062
retained groups.\ %
All three were authors but had not participated in benchmark construction.\ %
They assigned initial accept-or-reject labels based on query-target
consistency, target ambiguity, and context adequacy and diversity.\ %
Initial judgments yielded 1,598 unanimous acceptances and 297 unanimous
rejections.\ %
Of the 167 disputed groups, 127 received two accept votes and 40 received two
reject votes.\ %
Before discussion, three-way agreement was 91.90\% (1,895/2,062), and nominal
Krippendorff's $\alpha$ was 0.815.\ %
Disputed groups were adjudicated primarily through discussion.\ %
The final review accepted 1,692 groups and rejected 370.\ %

\textbf{Release.}
To improve the balance of task-specific displacement and rotation bins, we
subsampled accepted groups in Translation, Rotation, and Composition, omitting
191 groups.\ %
One additional ROG group was excluded after its context images triggered a
safety refusal from GPT-Image-2.\ %
The frozen benchmark contains 1,500 groups from 513 scenes across the three
source datasets.\ %

\input{sections/llm_review_prompt}
\flushbottom

\input{sections/task_examples}

\section{Evaluation Protocol Details}
\label{sec:metric-details}

This section details the six metrics in \Cref{sec:evaluation-method},
which assess agreement with the real target and the generation's intrinsic
quality.\ %

\input{sections/metric_protocol_details}

\FloatBarrier

\section{Training and Inference Details}
\label{sec:sft-details}

\subsection{Training Resource}
\label{sec:training-resource-record}
The released training resource contains 20,000 examples from 3,406 scenes,
each with six context images, a language query, and a real target image.\ %
Training examples are constructed from the training splits of all three
source datasets, while the benchmark uses their validation splits for held-out
evaluation.\ %
The training resource and benchmark share no scenes within any source dataset.\ %

\subsection{Released Training Configuration}
The release configuration specifies 10,000 optimizer updates with eight
distributed processes and no gradient accumulation.\ %
It uses AdamW \citeblue{adamw2019} with learning rate $10^{-5}$, cosine decay to $10^{-6}$,
200 warmup steps, gradient clipping at 0.5, and EMA decay 0.9995.\ %
The VAE and vision transformer are frozen, while the language model and
understanding branch remain trainable.\ %
Conditioning dropout is disabled.\ %
Images are center-cropped and resized to $512\times512$ for the VAE and
$518\times518$ for the vision transformer, with training timestep shift 4.\ %
These settings describe the released training recipe.\ %

\subsection{Generation Configuration}
The saved BAGEL-SFT inference configuration uses 50 sampling steps,
text and image guidance scales of 4 and 2, timestep shift 3,
and no-thinking inference.\ %
Input and saved output images are $512\times512$.\ %
Local generators share a deterministic per-sample seed across the
supplementary context settings.\ %
Closed-source APIs receive no seed.\ %
In our setup, directly supplying six separate images to Qwen-Image-Edit-2511
and Emu3.5 led to inference failures or severely degraded outputs.\ %
We therefore pack supporting views into a contact sheet and provide the anchor
separately when present.\ %
For anchor-free tasks, all context views are packed into a single contact sheet.\ %

\section{Spatial-Understanding Evaluation Details}
\label{sec:transfer-details}
All results in \Cref{tab:spatial-understanding} are obtained in our own
evaluations.\ %
SAT-v2 uses 150 questions evaluated in two answer orders.\ %
Micro accuracy counts correct predictions among all 300 responses,
while both-order accuracy counts questions answered correctly in both
orders among the 150 pairs.\ %
BAGEL and BAGEL-SFT use identical inputs and greedy, non-thinking decoding.\ %
Their correct counts are 174 and 188 responses, respectively,
and 41 and 62 complete pairs.\ %
BAGEL-SFT answers 322 of 1,000 MMSI-Bench questions correctly under the
Vanilla non-circular protocol and 383 of 1,050 MindCube-Tiny questions
under raw-QA evaluation.\ %
Our reproduced BAGEL accuracies are 31.0\% and 34.7\%, respectively,
matching SenseNova-SI \citeblue{exp_sensenova_si2026} at the reported precision.\ %
Eval.\ Size in \Cref{tab:spatial-understanding} denotes the number of
evaluated responses or question pairs under each protocol.\ %

\section{MindCue Implementation}
\label{sec:selection-details}
\input{sections/mindcue_implementation_details}

\FloatBarrier

% Keep the rigid result blocks adjacent without stretching section gaps.
\raggedbottom
\setlength{\intextsep}{2pt}
\setlength{\abovecaptionskip}{4pt}
\section{Supplementary Results}
\label{sec:supplementary-results}

This section complements the main experiments with component, context-scaling,
context-selection, and qualitative results on the same frozen 1,500 samples
and under the same evaluation protocols.\ %

\subsection{Task-Level Component Scores on MindView}
\Crefrange{tab:baseline-task-components-1}{tab:baseline-task-components-4}
report all six component scores for each task under the standard six-view input,
complementing the task composites in \Cref{tab:c6-results}.\ %
The comparison includes eleven off-the-shelf generators and BAGEL-SFT.\ %
BAGEL-SFT scores higher than BAGEL on all 42 task--component pairs and ranks
first among the twelve models on 31 of them.\ %
It leads Semantic, Spatial, and TScore on all seven tasks.\ %
Its Quality score also exceeds BAGEL on every task, although GPT-Image-2 remains
highest on that component throughout.\ %
Task abbreviations follow \Cref{sec:task-labels}.\ %
Bold and underline indicate the highest and second-highest scores across
all twelve models within each task.\ %
Sem., Spa., and Qual.\ denote Semantic, Spatial, and Quality.\ %
\textsuperscript{\ensuremath{\dagger}} denotes tiled context views, as in
\Cref{tab:final-c6-overall}.\ %

\input{tables/base_generators_c6_task_component_metrics}

\subsection{Context-Scaling Component Scores}
\Cref{tab:appendix-context-metrics} reports task-macro component scores
under the reduced $C_3$ and expanded $C_{10}$ settings.\ %
The corresponding standard $C_6$ scores appear in
\Cref{tab:final-c6-overall}.\ %
From $C_6$ to $C_{10}$, Semantic and Spatial decrease for all eleven
off-the-shelf generators, whereas Quality increases for nine.\ %
Over the same change, DINO-I and RScore decrease for ten generators, and
TScore decreases for nine.\ %
For the off-the-shelf generators, the common Overall decline therefore does
not imply that every component worsens as context expands.\ %
Bold and underline mark the highest and second-highest scores within
each context setting, and \textsuperscript{\ensuremath{\dagger}} again denotes
tiled context views.\ %

\input{tables/base_generators_context_component_metrics}

\subsection{Detailed Results for MindCue}
\Cref{tab:selection-component-metrics} decomposes the Overall results in
\Cref{tab:training_free_effectiveness} into six task-macro components for
the same five context policies on BAGEL and BAGEL-SFT.\ %
Fixed $C_6$ and Fixed $C_{10}$ are fixed-context baselines.\ %
RandDrop is a count-matched random control evaluated with one selection seed.\ %
It retains the same number of additions as MindCue for each sample.\ %
CounterSel instead retains the candidate additions rejected by MindCue and is
not count-matched.\ %
MindCue uses the query and predicted scene evidence to select additions
without updating either generator.\ %
Bold and underline mark the highest and second-highest policy scores for each
model and component.\ %
Against RandDrop, MindCue is higher on five of six components for BAGEL and all
six for BAGEL-SFT.\ %
Quality is the only BAGEL exception, by 0.3 points.\ %
The largest gains over RandDrop are in RScore, at 2.6 points for BAGEL and
1.6 points for BAGEL-SFT.\ %
Relative to Fixed $C_6$, MindCue is higher on all six components for BAGEL and
five for BAGEL-SFT, with only a 0.3-point decrease in Quality for BAGEL-SFT.\ %

\input{tables/training_free_context_selection_components}

\Needspace{8\baselineskip}
\subsection{Qualitative Comparisons}
\label{sec:qualitative-comparisons}
\begin{figure}[H]
    \centering
    \includegraphics[width=\linewidth]{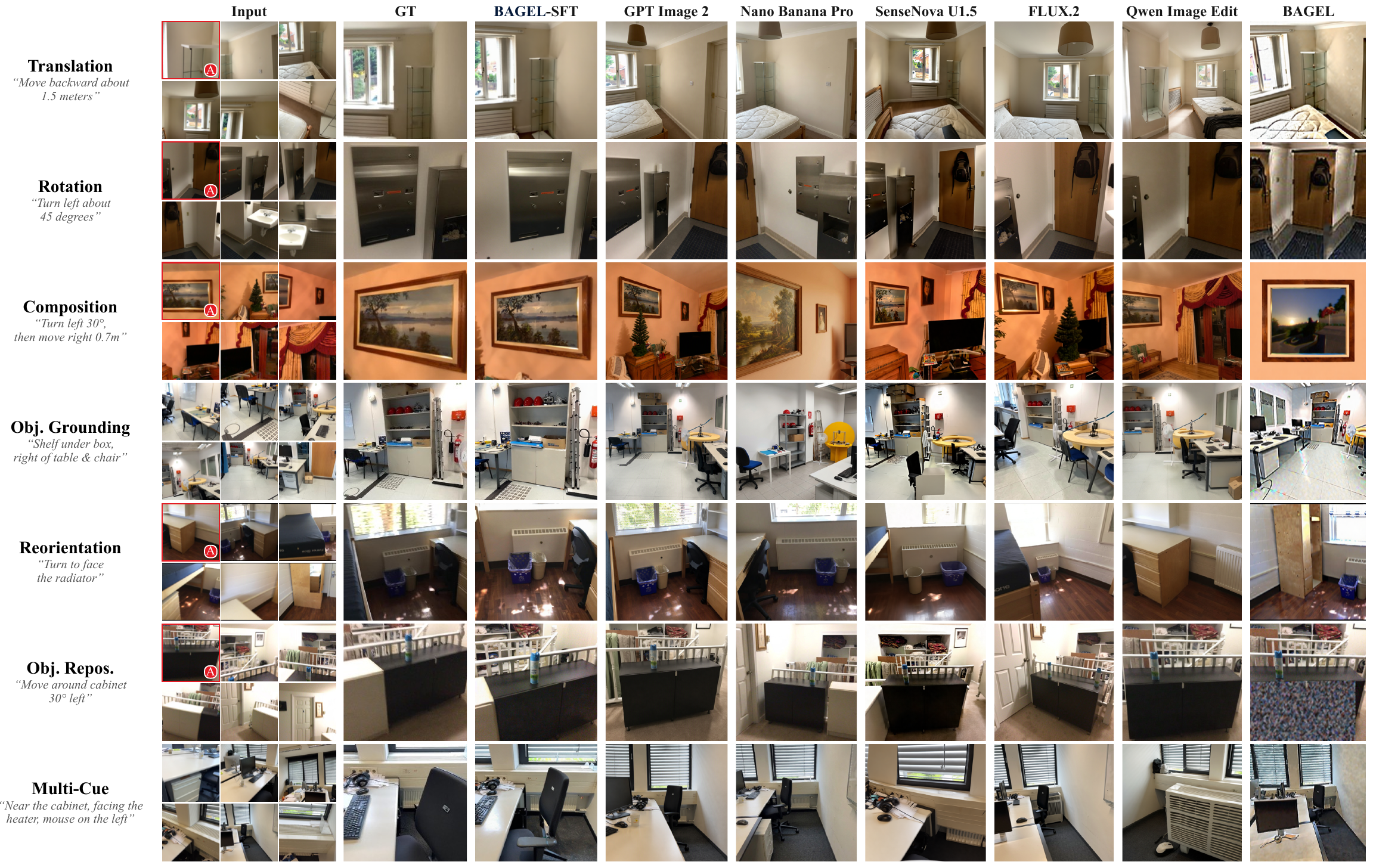}
    \caption{Qualitative comparison across all seven MindView tasks. Each row
    shows a query summary, six input contexts, the held-out real target (GT),
    and outputs from the selected generators named above the columns. A red
    border and \textbf{A} identify the anchor when one is required.}
    \label{fig:appendix-baseline-qualitative}
\end{figure}

\Cref{fig:appendix-baseline-qualitative} extends
\Cref{fig:baseline-qualitative} from two tasks to all seven.\ %
The examples support separate inspection of scene preservation and target-view
agreement across task types, complementing the component scores above.\ %

\clearpage
\input{sections/limitations}

%% file: sections/llm_review_prompt.tex
% Exact prompt used for GPT-5.6 candidate screening.
% Only LaTeX escaping and line/paragraph layout have been added.

\begin{tcolorbox}[
    breakable,
    sharp corners,
    colback=black!2,
    colframe=black!30,
    colbacktitle=black!6,
    coltitle=black,
    boxrule=0.4pt,
    titlerule=0.4pt,
    boxsep=0pt,
    left=8pt,
    right=8pt,
    top=6pt,
    bottom=6pt,
    toptitle=5pt,
    bottomtitle=5pt,
    before skip=8pt,
    after skip=8pt,
    fonttitle=\small\bfseries,
    fontupper=\small\rmfamily,
    title={GPT-5.6 screening prompt.},
    title after break={GPT-5.6 screening prompt (continued).},
    nameref={GPT-5.6 screening prompt},
    phantomlabel={sec:llm-review-prompt}
]
\setlength{\parindent}{0pt}
\setlength{\parskip}{4pt}
You are screening one MindView candidate sample.\par
{\raggedright
\textbf{Inputs:}
\mbox{- TASK: \{task\_type\}}\quad
\mbox{- QUERY: \{query\}}\quad
\mbox{- Six images labeled CONTEXT 1--6}\quad
\mbox{- One held-out image labeled TARGET}\quad
\mbox{- ANCHOR: \{1 or NONE\}}\par}
\textbf{Return PASS only if:}
\begin{enumerate}[label=\arabic*.,leftmargin=1.4em,
    topsep=0pt,itemsep=4pt,parsep=0pt,partopsep=0pt]
\item The QUERY and CONTEXT images uniquely constrain the target camera position
and viewing direction. Return FAIL if objects or spatial cues are ambiguous,
or if substantially different viewpoints could equally satisfy the QUERY or
produce similarly plausible observations due to repeated structures, scene
symmetry, or insufficient cross-view evidence.
\item The six CONTEXT images provide complementary scene information and are not
dominated by duplicate or near-duplicate viewpoints.
\end{enumerate}
Use TARGET only to check consistency, not to resolve ambiguity. Use only
visible evidence. If uncertain, return FAIL.
\textbf{Return exactly one word:} PASS or FAIL
\end{tcolorbox}

%% file: sections/task_examples.tex
\subsection{Task Examples and Generation Prompts}
\label{sec:task-examples}

\Crefrange{fig:task-example-translation}{fig:task-example-mcvi}
show one released example for each task in the order introduced in
\Cref{sec:task-labels}.\ %
Each example contains six context images ordered from left to right,
the held-out ground-truth target (GT), and the full generation prompt.\ %
A red border and ``A'' mark the first image when an anchor is required.\ %
The target is shown for reference and is not a generation input.\ %

% Preserve the supplied PDFs at full text width and balance the seven
% examples across three pages.
\begingroup
\setlength{\intextsep}{2pt}
\setlength{\abovecaptionskip}{4pt}

\begin{figure}[H]
    \centering
    \includegraphics[width=\textwidth]{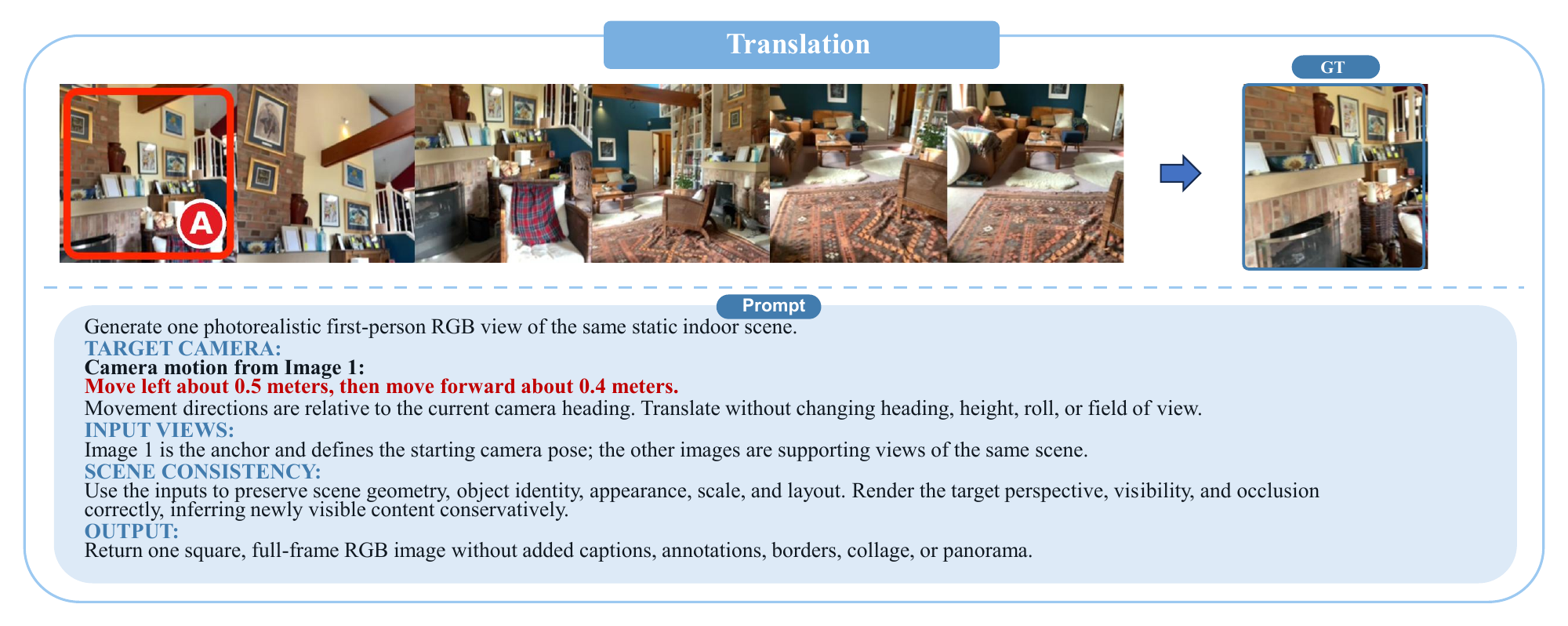}
    \caption{Translation example from ARKitScenes.}
    \label{fig:task-example-translation}
\end{figure}

\begin{figure}[H]
    \centering
    \includegraphics[width=\textwidth]{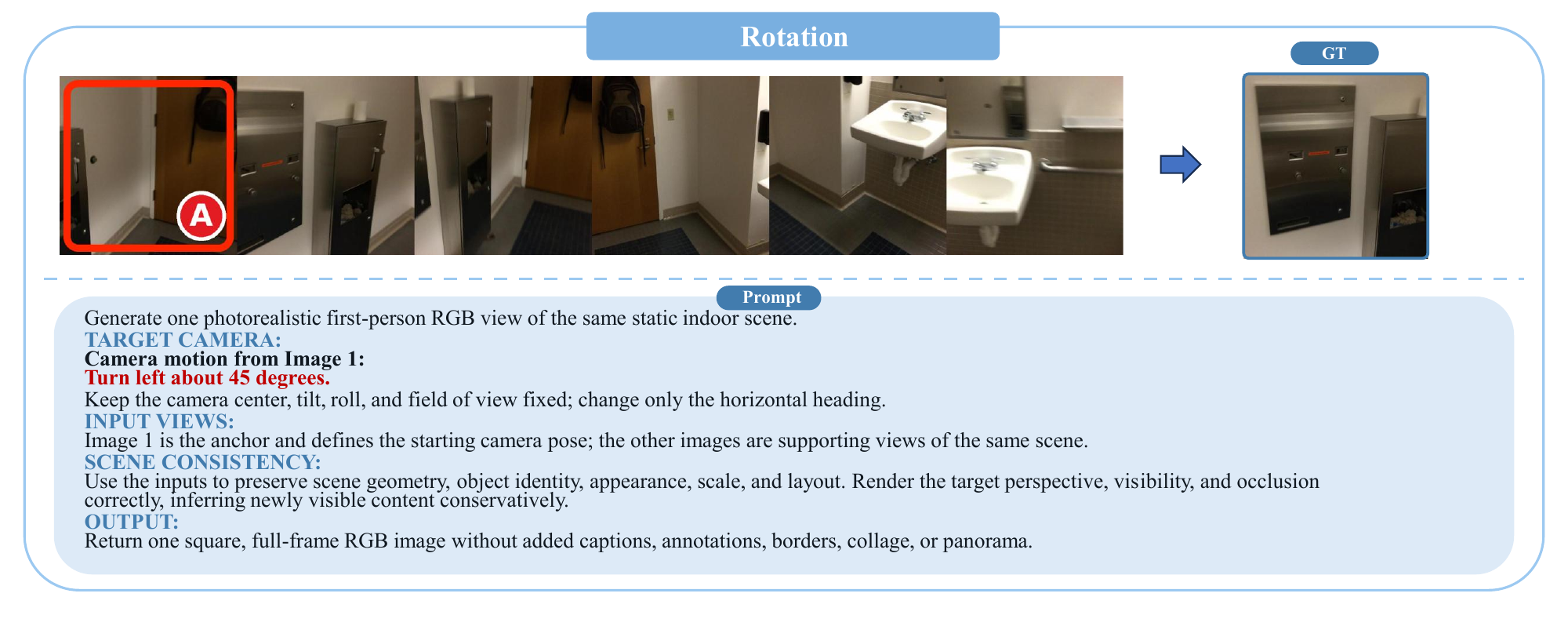}
    \caption{Rotation example from ScanNet.}
    \label{fig:task-example-rotation}
\end{figure}

\clearpage

\begin{figure}[H]
    \centering
    \includegraphics[width=\textwidth]{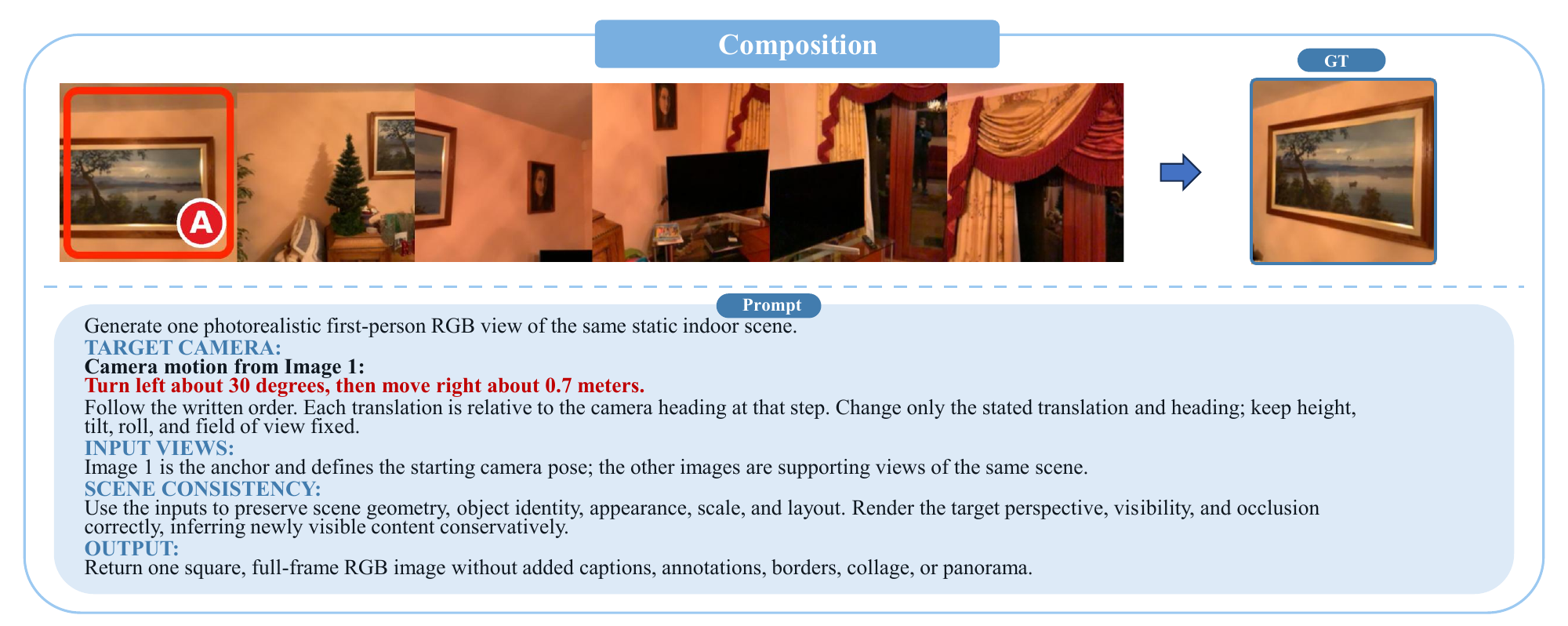}
    \caption{Composition example from ARKitScenes.}
    \label{fig:task-example-composition}
\end{figure}

\begin{figure}[H]
    \centering
    \includegraphics[width=\textwidth]{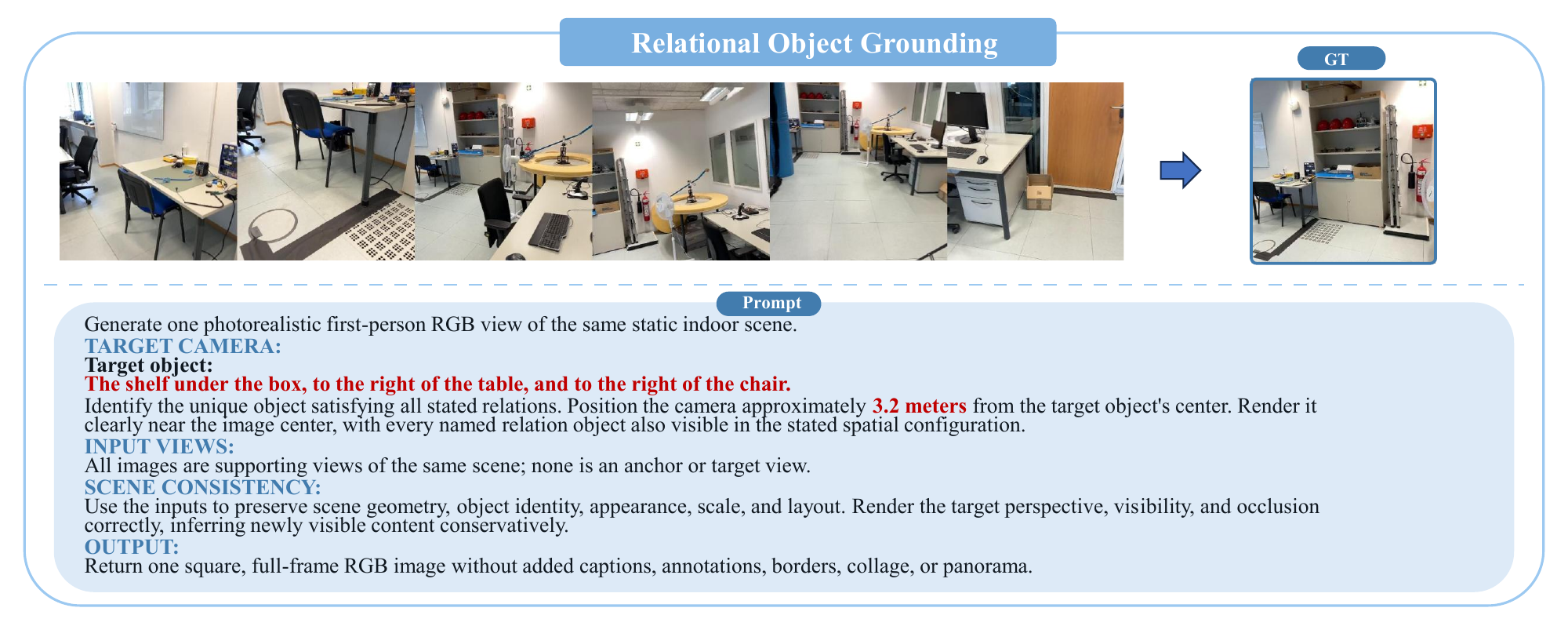}
    \caption{Relational Object Grounding example from ScanNet++.}
    \label{fig:task-example-rog}
\end{figure}

\begin{figure}[H]
    \centering
    \includegraphics[width=\textwidth]{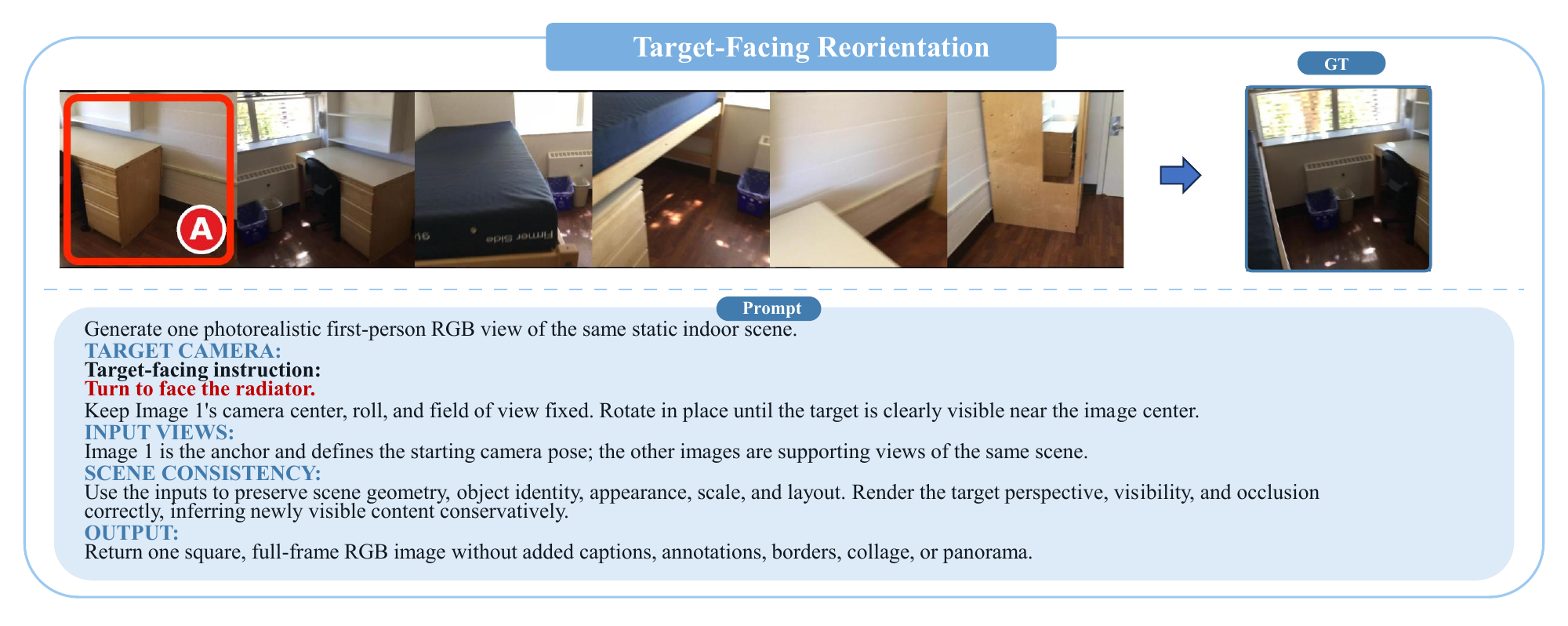}
    \caption{Target-Facing Reorientation example from ScanNet.}
    \label{fig:task-example-tfr}
\end{figure}

\clearpage

\begin{figure}[H]
    \centering
    \includegraphics[width=\textwidth]{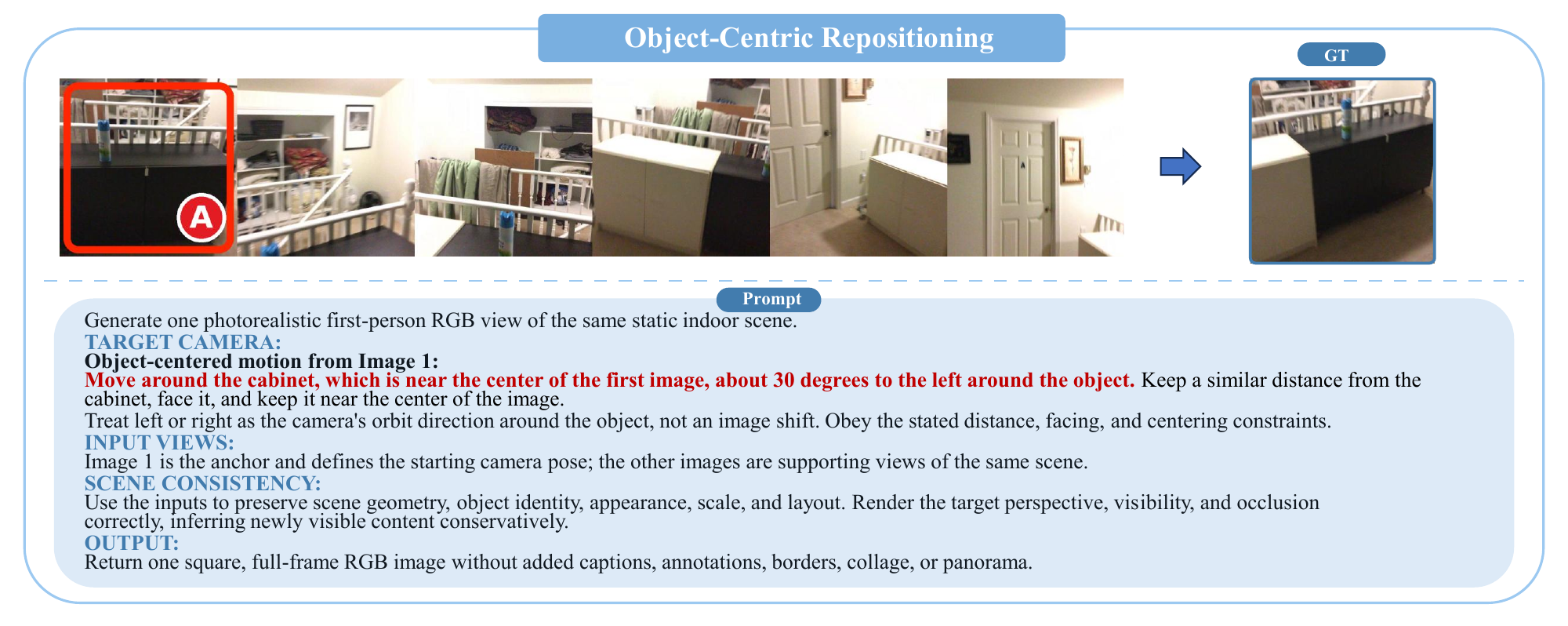}
    \caption{Object-Centric Repositioning example from ScanNet.}
    \label{fig:task-example-ocr}
\end{figure}

\begin{figure}[H]
    \centering
    \includegraphics[width=\textwidth]{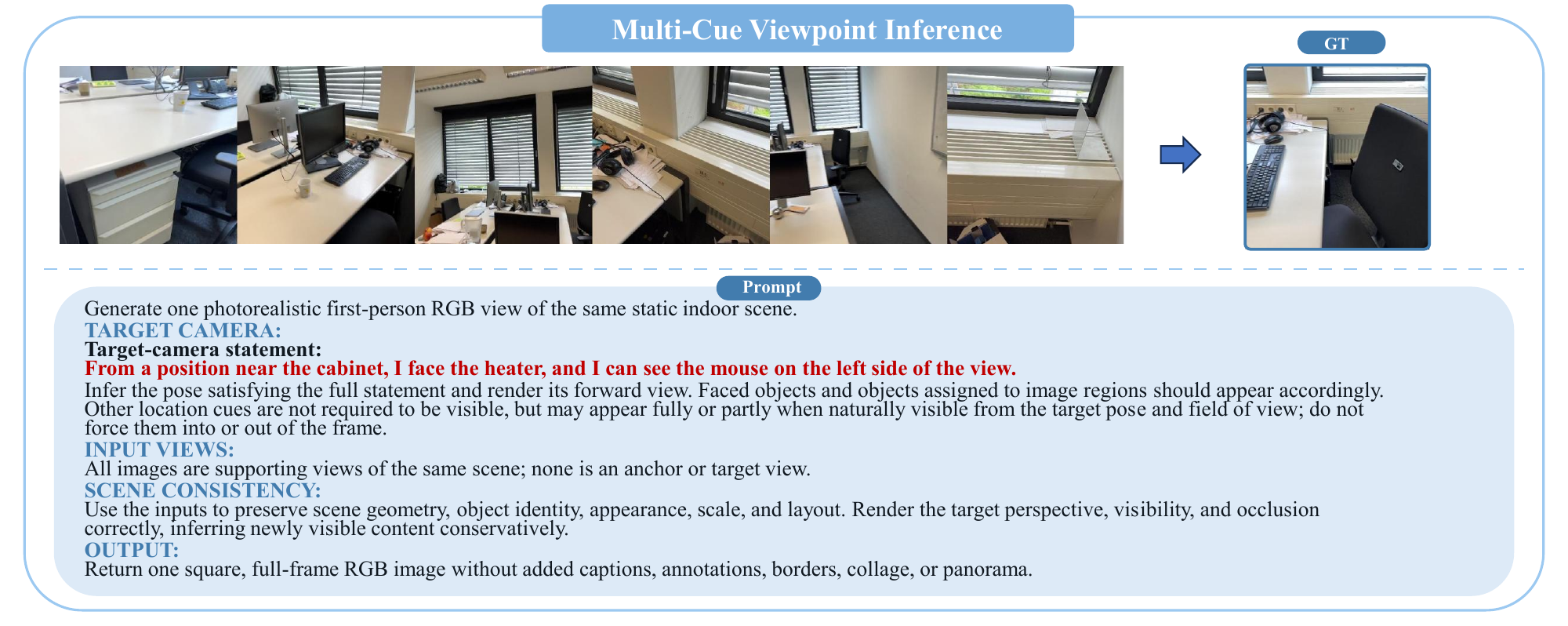}
    \caption{Multi-Cue Viewpoint Inference example from ScanNet++.}
    \label{fig:task-example-mcvi}
\end{figure}
\endgroup

% Keep the final two examples together and let Appendix B continue below them.

%% file: sections/metric_protocol_details.tex
\subsection{Image-Feature Similarity}
DINO-I compares the visual content of a generated image with its real target
using the pretrained DINOv2-Large encoder \citeblue{dinov22023}.\ %
Let $\mathbf f_i^{\mathrm{gen}}$ and $\mathbf f_i^{\mathrm{real}}$ denote
their unit-normalized final-layer CLS embeddings.\ %
The score is their cosine similarity scaled by 100,
\begin{equation}
\mathrm{DINO\text{-}I}_i
 =100\,(\mathbf f_i^{\mathrm{gen}})^\top\mathbf f_i^{\mathrm{real}}.
\label{eq:appendix-dino-score}
\end{equation}
No clipping or offset is applied, so negative similarities remain negative.\ %
Before feature extraction, both images are converted to RGB, center-cropped
to a square, and resized to $512\times512$ with Lanczos interpolation.\ %
The encoder's standard preprocessing then produces normalized
$224\times224$ inputs.\ %

\subsection{VLM-Based Assessment}
\label{sec:judge-protocol}
We use Qwen3-VL-30B-A3B-Instruct \citeblue{qwen3vl_judge_model} for two
separate evaluations.\ %
The first receives the real target followed by the generation, labeled as
the reference and candidate images, respectively.\ %
It returns separate Semantic and Spatial scores for scene-specific content
and agreement in viewpoint and layout.\ %
The second receives only the generation and assigns a Quality score based
on its internal coherence and visible defects.\ %

Neither evaluation receives the query, context views, task type, camera
metadata, or the other evaluation's response.\ %
Spatial therefore measures agreement with the target image, not direct
compliance with the language query.\ %
The first evaluation uses the initial RGB conversion, square crop, and
$512\times512$ resize described above.\ %
For Quality, the full RGB image is reduced to fit within $512\times512$
without upscaling, preserving its aspect ratio, borders, and peripheral
defects.\ %
The judge's standard image preprocessing follows in both cases.\ %

Each evaluation uses deterministic decoding with at most 256 output tokens.\ %
The three ratings must be integers from 0 to 10 and are multiplied by ten
for reporting.\ %
Missing or invalid ratings constitute evaluation failures rather than zero
scores.\ %
\Cref{sec:evaluator-prompts} reproduces both prompts verbatim,
apart from typesetting and line wrapping.\ %

\subsection{Geometric Consistency}
\label{sec:vggt-metric-protocol}
We use the pretrained VGGT-1B model \citeblue{vggt2025} to measure agreement
between the generated and real target viewpoints.\ %
For each sample, VGGT jointly processes ten ordered reference views, the real
target, and the generation, in that order.\ %
The references are drawn from the sample's ten-context set and held fixed
across generators and context budgets.\ %
Images undergo the same initial square preprocessing and are then resized
to $518\times518$ for VGGT.\ %

\textbf{Camera-pose comparison.}
Joint inference places the reference views and both target images in a shared
reconstructed coordinate system for each evaluation.\ %
For sample $i$, $k\in\{\mathrm{gen},\mathrm{real}\}$ indexes the generated
and real target images.\ %
Let $\mathbf x_w\in\mathbb R^3$ denote a point's coordinates in this world
frame and $\mathbf x_c\in\mathbb R^3$ its coordinates in image $k$'s camera
frame.\ %
The estimated rotation $\hat R_{i,k}\in\mathrm{SO}(3)$ and translation
$\hat{\mathbf t}_{i,k}\in\mathbb R^3$ map world coordinates to camera
coordinates through
$\mathbf x_c=\hat R_{i,k}\mathbf x_w+\hat{\mathbf t}_{i,k}$.\ %
Setting $\mathbf x_c=\mathbf0$ gives the camera center in the world frame,
$\hat{\mathbf c}_{i,k}=-\hat R_{i,k}^{\top}\hat{\mathbf t}_{i,k}$.\ %
We compare camera centers, rather than extrinsic translation vectors, and
compute the errors in \Cref{eq:target-view-pose-errors} as
\begin{equation}
\begin{aligned}
e_{\mathrm{rot},i}
 &=\frac{180}{\pi}\arccos\!\left(
 \operatorname{clip}\!\left(
 \frac{\operatorname{tr}(\hat R_{i,\mathrm{gen}}
                        \hat R_{i,\mathrm{real}}^\top)-1}{2},-1,1
 \right)\right),\\
e_{\mathrm{trans},i}
 &=a_i\left\|\hat{\mathbf c}_{i,\mathrm{gen}}
            -\hat{\mathbf c}_{i,\mathrm{real}}\right\|_2.
\end{aligned}
\label{eq:vggt-metric-errors}
\end{equation}
Here, $\operatorname{tr}$ denotes the matrix trace, $\|\cdot\|_2$ the
Euclidean norm, and $\operatorname{clip}(x,l,u)$ restricts $x$ to $[l,u]$.\ %
The rotation error is in degrees.\ %
The positive scale $a_i$ in \Cref{eq:vggt-depth-scale} converts
reconstructed lengths to meters, so $e_{\mathrm{trans},i}$ measures
camera-center displacement in meters.\ %

\textbf{Depth-based scale calibration.}
We estimate $a_i$ from sensor depths for the ten reference views and the real
target, denoted collectively by $\mathcal J_i$.\ %
Measured depths are converted to meters, and invalid readings are excluded.\ %
Nearest-neighbor resampling aligns them with the predicted depth maps under
the same centered image crop.\ %
For each real frame, we first retain pixels with finite measured and
predicted depths, positive measured depth, and predicted depth above
$10^{-8}$.\ %
When at least 512 such pixels have finite depth confidence, we select those
at or above their median confidence, provided that at least 256 remain.\ %
Otherwise, we retain the original valid set.\ %

Let $\mathcal V_{if}$ be the retained pixels in frame $f$, with measured
depth $d^{\mathrm{sens}}_{ifp}$ and predicted depth $\hat d_{ifp}$ at pixel $p$.\ %
We aggregate log-depth ratios first within each frame and then across frames,
\begin{equation}
a_i=\exp\!\left[
\operatorname{median}_{f\in\mathcal J_i}
\left\{
\operatorname{median}_{p\in\mathcal V_{if}}
\left(\log d^{\mathrm{sens}}_{ifp}-\log\hat d_{ifp}\right)
\right\}\right].
\label{eq:vggt-depth-scale}
\end{equation}
Calibration requires at least 256 retained pixels per real frame and a
finite, positive $a_i$.\ %
The generation participates in joint inference but contributes no sensor
depth to calibration.\ %
The real target and sensor depths are used only for evaluation, and neither
pose error requires ground-truth camera poses.\ %

\textbf{Geometric scores.}
We convert the errors to higher-is-better scores before averaging samples,
\begin{equation}
\mathrm{RScore}_i=100\,\operatorname{clip}\!\left(
1-\frac{e_{\mathrm{rot},i}}{60^\circ},0,1\right),\qquad
\mathrm{TScore}_i=100\,\operatorname{clip}\!\left(
1-\frac{e_{\mathrm{trans},i}}{2\,\mathrm{m}},0,1\right).
\label{eq:experiment-geometry-scores}
\end{equation}
Each score decreases linearly from 100 at zero error to zero at its threshold
and remains zero for larger errors.\ %

\subsection{Evaluator Reliability}
\label{sec:evaluator-reliability}
% The rigid H floats in this subsection can leave a short page before the
% next table.  Avoid stretching the subsection-title gap to fill that page.
\raggedbottom

\textbf{Human--VLM ranking agreement.}
We assess whether the VLM judge reflects human preferences on a score-blind
subset of 210 standard-six-view samples, with 30 samples from each task.\ %
The same three multimodal-research experts who conducted the benchmark review
in \Cref{sec:context-review-release} independently rate outputs from
the four models in \Cref{tab:human-vlm-agreement} using the same Semantic,
Spatial, and Quality criteria as the judge.\ %
The table reports the expert-averaged human scores and VLM scores on the same
subset.\ %

\input{tables/human_vlm_agreement}

For 194 of the 210 samples (92.4\%), all three experts independently give the
same four-model ranking, and this unanimous human ranking exactly matches the
ranking produced by the VLM judge.\ %
Absolute scores nevertheless reveal a consistent calibration shift.\ %
For every model, the VLM judge assigns lower Semantic and Spatial scores and
higher Quality scores than the experts.\ %
Despite this shift, all six aggregate pairwise model comparisons agree in
direction for each component (18/18 overall), yielding the same complete
ordering: GPT-Image-2, Nano Banana Pro, SenseNova-U1.5, and BAGEL.\ %
These results support the judge's comparative validity on the evaluated
subset despite differences in absolute calibration.\ %

\textbf{Reliability of VGGT-based geometric scores.}
Joint reconstruction introduces a different concern: a degraded generation
could perturb the real-view geometry against which it is scored.\ %
We evaluate this sensitivity quantitatively and inspect score behavior in
a severe qualitative case.\ %
We sample 30 cases uniformly without replacement per task,
covering 210 samples from 150 scenes.\ %
For each, we substitute the standard-six-view outputs of BAGEL, BAGEL-SFT,
FLUX.2, and GPT-Image-2, fixing the ten reference images, real target (GT),
input order, and preprocessing.\ %
Each reconstruction is independently calibrated to meters from the eleven
real views' sensor depths (\Cref{eq:vggt-depth-scale}).\ %
As a repeatability control, identical BAGEL inputs are rerun on three
samples per task; all 21 reruns yield exactly identical saved poses,
intrinsics, depths, confidence maps, and calibration scales.\ %

We compare the fixed real views across generator conditions.\ %
For the GT, $\Delta R$ averages changes in its relative rotations to the
ten reference cameras, and $\Delta C$ measures camera-center displacement
after rigid alignment fitted only to those reference centers, without
rescaling.\ %
For the references, $\Delta R$ and $\Delta L$ average relative-rotation
changes and absolute camera-distance changes over all 45 camera pairs.\ %
These diagnostics remove global-coordinate ambiguity while retaining
differences in calibrated scale.\ %
For each real image, depth change is
$\Delta D=100\,\operatorname{median}_{p}
\bigl[2|d_p^{(a)}-d_p^{(b)}|/(d_p^{(a)}+d_p^{(b)})\bigr]$,
where $a,b$ index generator conditions, using calibrated depths on the full
$518\times518$ grid and a fixed valid-sensor-depth mask across conditions.\ %
We report GT depth change separately and average the ten per-image changes
for the references.\ %
\Cref{tab:vggt-reference-stability} reports medians across samples
after averaging each diagnostic over the six unordered generator pairs
within each sample; the identical-input reruns are excluded.\ %
The pooled median changes are $0.71^\circ$, 2.75\,cm, and 0.76\% for GT
rotation, position, and depth; reference changes are $0.75^\circ$,
2.63\,cm in pairwise distance, and 0.87\% in depth.\ %
These small median changes indicate that outputs from different generators
typically have little effect on the reconstructed geometry of the fixed
real views, supporting the stability of VGGT-based geometric evaluation.\ %
\input{tables/vggt_reference_stability}

\begin{figure}[H]
    \centering
    \includegraphics[width=\linewidth]{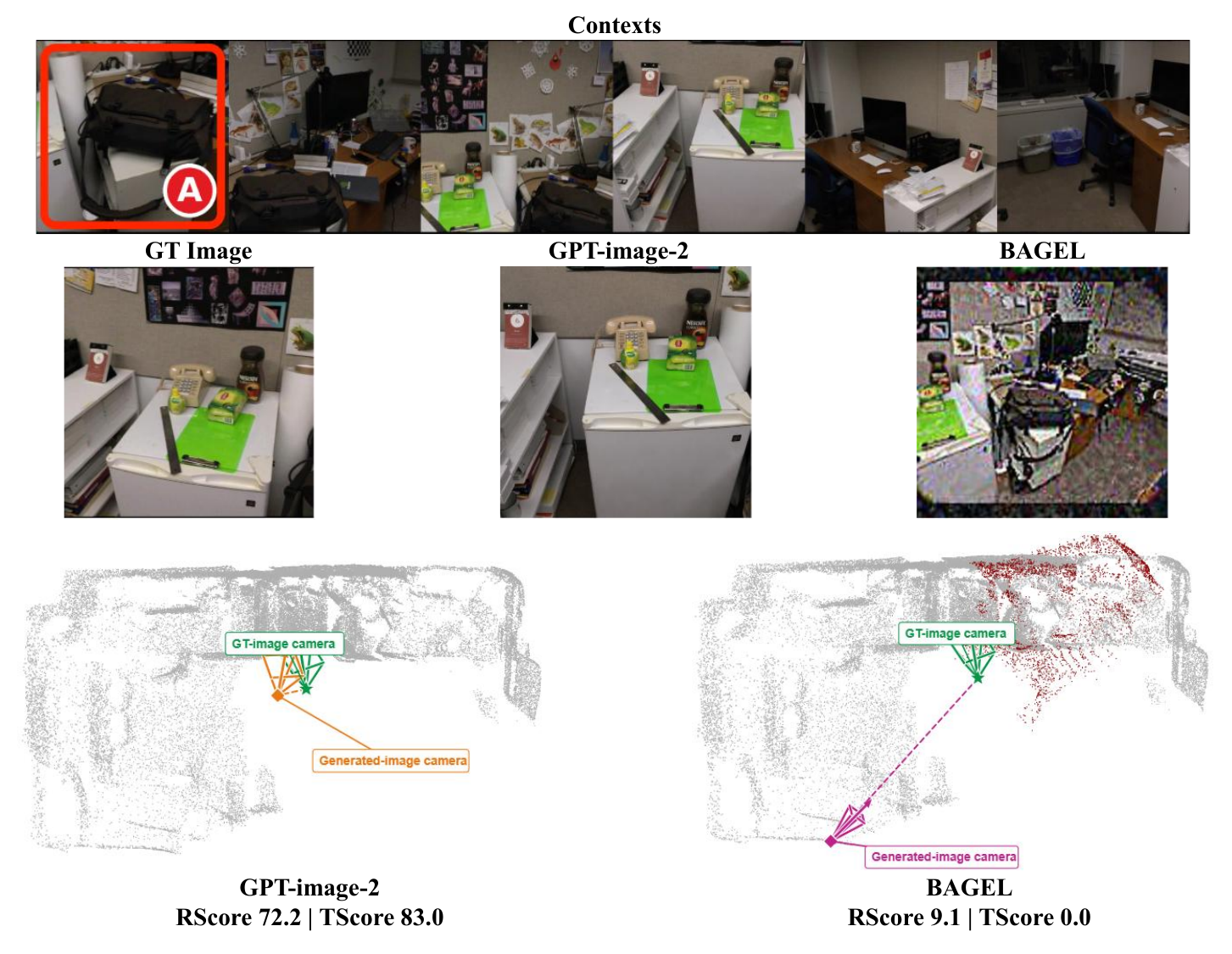}
    \caption{\textbf{VGGT evaluation of GPT-Image-2 and BAGEL outputs.}
    For the ScanNet TFR query ``Turn to face the telephone,'' the top row
    shows six generation contexts (\textbf{A}: anchor); evaluation uses ten
    fixed references, the real target, and one generation.
    Reconstructions are independently depth-calibrated and rigidly aligned
    using only the ten reference camera centers, without scale fitting.
    All cameras are VGGT estimates; GT denotes the real target image.
    Dark red marks points contributed by BAGEL's generation;
    all other scene points are gray.}
    \label{fig:vggt-candidate-substitution}
\end{figure}

\Cref{fig:vggt-candidate-substitution} complements the aggregate
analysis with a visibly degraded generation.\ %
GPT-Image-2 preserves the target-facing scene content, whereas BAGEL
introduces severe appearance and layout distortions.\ %
Their RScore/TScore values are 72.2/83.0 and 9.1/0.0, respectively.\ %
The reference scene and estimated GT camera remain visually similar after
alignment, while BAGEL's estimated camera lies much farther from the
estimated target camera.\ %
This case illustrates score behavior consistent with the visible
target-view discrepancy despite the degraded generation.\ %

Together, the quantitative and qualitative results support RScore and
TScore as meaningful measures of target-view geometric consistency.\ %
Reference geometry exhibits small median changes across model outputs.\ %
In the illustrated case, it remains visually stable despite severe
generation artifacts, while the scores reflect the visible viewpoint
discrepancy.\ %

\FloatBarrier
\flushbottom

\subsection{Score Aggregation}
For each generator and context setting, let $s_{itj}$ be sample $i$'s score
on metric $j$ for task $t$ after the conversions above, and let $n_t$ be the
number of samples in that task.\ %
We first average samples within each task, then give equal weight to the six
components and the seven tasks,
\begin{equation}
\bar s_{tj}=\frac{1}{n_t}\sum_{i=1}^{n_t}s_{itj},\qquad
S_t=\frac{1}{6}\sum_{j=1}^{6}\bar s_{tj},\qquad
\mathrm{Overall}=\frac{1}{7}\sum_{t=1}^{7}S_t.
\label{eq:experiment-aggregation}
\end{equation}
Each benchmark-level component score is likewise averaged equally across
the seven tasks, so tasks with more samples do not receive greater weight.\ %

Aggregation requires valid scores for every benchmark sample.\ %
Failed or missing evaluations are neither assigned zero nor silently
omitted.\ %
Scores and gains are displayed to one decimal place, while differences
and rankings are computed before rounding.\ %
Overall is a composite score, not a calibrated accuracy.\ %
Comparisons are descriptive, without significance tests.\ %

\Needspace{7\baselineskip}
\subsection{Evaluator Prompts}
\label{sec:evaluator-prompts}
\input{sections/judge_prompts}

%% file: tables/human_vlm_agreement.tex
\begingroup
\setlength{\intextsep}{7pt}
\begin{table}[H]
\centering
\caption{\textbf{Human and VLM-judge component scores on the shared 210-sample
subset.} Human scores are three-expert averages; all values are on a 0--100
scale, and higher is better.}
\label{tab:human-vlm-agreement}
\begingroup
\fontsize{8}{9}\selectfont
\setlength{\tabcolsep}{6.5pt}
\renewcommand{\arraystretch}{1.04}
\begin{tabular}{@{}lrrrrrr@{}}
\toprule
\multirow{2}{*}{Model}
& \multicolumn{2}{c}{Semantic}
& \multicolumn{2}{c}{Spatial}
& \multicolumn{2}{c}{Quality} \\
\cmidrule(lr){2-3}\cmidrule(lr){4-5}\cmidrule(lr){6-7}
& Human & VLM & Human & VLM & Human & VLM \\
\midrule
GPT-Image-2      & 68.4 & 63.4 & 63.3 & 50.2 & 76.3 & 84.8 \\
Nano Banana Pro & 62.6 & 59.5 & 59.5 & 47.4 & 72.5 & 79.8 \\
SenseNova-U1.5  & 57.6 & 54.1 & 51.5 & 43.5 & 66.8 & 70.3 \\
BAGEL            & 45.5 & 45.4 & 43.0 & 33.4 & 44.1 & 48.6 \\
\bottomrule
\end{tabular}
\endgroup
\end{table}
\endgroup

%% file: tables/vggt_reference_stability.tex
\begin{table}[H]
\centering
\caption{\textbf{VGGT reference stability across generators.}
Only the generation changes; ten contexts and the real target stay fixed.
Entries are sample medians after averaging six generator pairs per sample.
$\Delta R$: relative-rotation change; $\Delta C$: aligned GT-center
displacement; $\Delta L$: context-pair distance change;
$\Delta D$: symmetric relative depth change.
Distances are depth-calibrated; lower values indicate greater stability.}
\label{tab:vggt-reference-stability}
\begingroup
\fontsize{8}{9}\selectfont
\setlength{\tabcolsep}{3.5pt}
\renewcommand{\arraystretch}{1.06}
\begin{tabular*}{\linewidth}{@{\extracolsep{\fill}}lr rrr rrr@{}}
\toprule
\multirow{2}{*}{Task} & \multirow{2}{*}{$N$}
& \multicolumn{3}{c}{Real target (GT)}
& \multicolumn{3}{c}{Contexts} \\
\cmidrule(lr){3-5}\cmidrule(l){6-8}
& & $\Delta R$ ($^\circ$) & $\Delta C$ (cm) & $\Delta D$ (\%)
  & $\Delta R$ ($^\circ$) & $\Delta L$ (cm) & $\Delta D$ (\%) \\
\midrule
Translation & 30 & 0.85 & 4.96 & 1.09 & 1.09 & 6.02 & 1.39 \\
Rotation & 30 & 0.52 & 1.92 & 0.58 & 0.65 & 2.14 & 0.79 \\
Composition & 30 & 0.96 & 4.60 & 0.95 & 0.88 & 4.71 & 1.18 \\
ROG & 30 & 0.82 & 3.25 & 0.99 & 0.93 & 2.62 & 0.97 \\
TFR & 30 & 0.63 & 1.60 & 0.55 & 0.70 & 1.88 & 0.80 \\
OCR & 30 & 0.50 & 1.96 & 0.48 & 0.54 & 2.04 & 0.67 \\
MCVI & 30 & 0.74 & 2.84 & 0.75 & 0.77 & 2.33 & 0.77 \\
\midrule
All sampled cases & 210 & 0.71 & 2.75 & 0.76 & 0.75 & 2.63 & 0.87 \\
\bottomrule
\end{tabular*}
\endgroup
\end{table}

%% file: sections/judge_prompts.tex
% Exact prompt text extracted from qwen_vl.py (qwen_judge_v2).
% Only LaTeX escaping, emphasis, and whitespace layout have been added.
% Prompts may continue across pages; do not abbreviate the release prompts.

\begingroup
\brokenpenalty=10000
\clubpenalty=10000
\widowpenalty=10000
\newcommand{\judgejson}[1]{\texttt{\fontsize{8}{9.5}\selectfont #1}}
\tcbset{mindview judge prompt/.style={
    breakable,
    sharp corners,
    colback=black!2,
    colframe=black!30,
    colbacktitle=black!6,
    coltitle=black,
    boxrule=0.4pt,
    titlerule=0.4pt,
    boxsep=0pt,
    left=8pt,
    right=8pt,
    top=6pt,
    bottom=6pt,
    toptitle=5pt,
    bottomtitle=5pt,
    before skip=8pt,
    after skip=8pt,
    fonttitle=\small\bfseries,
    fontupper=\fontencoding{T1}\fontsize{8.5}{9.5}\selectfont\rmfamily
}}

\begin{tcolorbox}[mindview judge prompt,
    title={Semantic and Spatial prompt.},
    title after break={Semantic and Spatial prompt (continued).},
    nameref={Semantic and Spatial prompt},
    phantomlabel={sec:judge-fidelity-prompt}]
\setlength{\parindent}{0pt}
\setlength{\parskip}{2.5pt}
% BEGIN EXACT PROMPT: build_fidelity_prompt
You are a strict, task-independent evaluator of image fidelity.
The labeled REFERENCE IMAGE is the ground-truth target.
The labeled CANDIDATE IMAGE is the model prediction.\par
\textbf{Important rules:}
- The two images may or may not depict the same scene. Do not assume that they match.
- Inspect each image independently before comparing them.
- Use only visible pixel evidence.
- Never attribute an object, region, or property to one image merely because it appears in the other.
- Do not use or infer any hidden prompt, context images, task type, or camera metadata.
- Evaluate Semantic Fidelity and Spatial Fidelity independently.
- Do not judge aesthetic preference or technical image quality in this evaluation.\par
\textbf{1. Semantic Fidelity: WHAT is present}\par
\textbf{Evaluate:}
- scene identity and scene category;
- main objects, structures, and recognizable content;
- approximate object count and identity-defining attributes;
- missing, extra, substituted, hallucinated, or unrecognizable content.\par
\textbf{Do NOT evaluate:}
- camera viewpoint, orientation, framing, crop, or visible side by itself;
- object position, scale, depth, occlusion, or relative layout;
- global color tone, lighting, rendering style, sharpness, or visual polish.\par
A spatially displaced object can still be semantically correct.
A viewpoint difference alone must not reduce Semantic Fidelity if the same important content remains visible and recognizable.
However, important content that is visibly missing or hallucinated is a semantic error, regardless of its cause.\par
\textbf{Semantic Fidelity scoring:}
10 = The same scene-specific content is preserved almost completely. All major objects and structures are present and recognizable, with no meaningful missing, extra, substituted, or hallucinated content.
8 = The same scene and all major content are preserved, with only minor errors involving secondary objects or identity-defining attributes.
6 = The core scene is clearly recognizable, but one major object or several secondary elements are missing, substituted, hallucinated, or unrecognizable.
4 = Only partial scene-specific content is preserved. Multiple major objects or structures are incorrect, although some meaningful correspondence remains.
2 = The images share only a generic scene category or a few isolated object categories; scene-specific correspondence is very weak.
0 = No meaningful semantic correspondence, or the candidate contains no interpretable scene content.\par
\textbf{Semantic score constraints:}
- Generic category similarity alone, such as both images being bedrooms or offices, cannot receive a score above 3.
- If one major or scene-defining object is missing, replaced, or hallucinated, the score cannot exceed 6.
- If multiple major objects are incorrect, the score cannot exceed 4.
- A score of 9 or 10 requires nearly complete scene-specific correspondence, not merely a similar category or overall impression.\par
\textbf{2. Spatial Fidelity: WHERE things are and FROM WHERE the scene is viewed}\par
\textbf{Evaluate:}
- camera viewpoint and orientation;
- framing, crop, and visible scene regions or surfaces;
- relative object positions, scale, depth ordering, and occlusion;
- scene layout, geometry, and composition.\par
Judge spatial relations using scene elements that can be meaningfully corresponded between the two images.\par
\textbf{Do NOT evaluate:}
- object category or semantic identity by itself;
- color, lighting, texture, rendering style, sharpness, or visual attractiveness;
- isolated generation artifacts unless they make spatial relationships impossible to assess.\par
Do not reward a generic similar composition unless corresponding scene elements support it.
If meaningful spatial correspondence cannot be established, use a score of 0, 1, or 2.\par
\textbf{Spatial Fidelity scoring:}
10 = Viewpoint, orientation, framing, visible regions, and spatial layout essentially match, with only negligible offsets.
8 = The intended view and layout are correct, with only minor shifts, crop or scale differences, or small geometric errors.
6 = The overall view is similar, but noticeable differences exist in camera pose, framing, visibility, scale, or relative placement.
4 = Some spatial correspondence exists, but the viewpoint, visible region, or layout is clearly incorrect.
2 = Very little spatial correspondence remains; there is a major camera-pose, framing, or layout mismatch.
0 = No usable spatial correspondence, or spatial structure cannot be assessed.\par
\textbf{Spatial score constraints:}
- A clearly incorrect viewing direction or target-facing direction cannot receive a score above 4.
- A major mismatch in the visible scene region or room side cannot receive a score above 4.
- Generic centered composition without scene-specific correspondence cannot receive a score above 2.
- A score of 9 or 10 requires both camera view and scene layout to match closely.\par
Use integers from 0 to 10. Integers between the stated anchors may be used when justified.
Apply all score constraints before producing the final scores.\par
Each reason must identify the strongest visible evidence and explicitly distinguish the Reference from the Candidate.
Keep each reason to one short sentence.\par
{\raggedright
Return ONLY valid JSON in exactly this format:
\judgejson{\{ "semantic\_score": \textless{}integer 0-10\textgreater{}, "spatial\_score": \textless{}integer 0-10\textgreater{}, "semantic\_reason": "\textless{}short evidence-based reason\textgreater{}", "spatial\_reason": "\textless{}short evidence-based reason\textgreater{}" \}}\par}
% END EXACT PROMPT: build_fidelity_prompt
\end{tcolorbox}

\begin{tcolorbox}[mindview judge prompt,
    title={Quality prompt.},
    title after break={Quality prompt (continued).},
    nameref={Quality prompt},
    phantomlabel={sec:judge-quality-prompt}]
\setlength{\parindent}{0pt}
\setlength{\parskip}{2.5pt}
% BEGIN EXACT PROMPT: build_quality_prompt
You are a strict evaluator of the intrinsic visual quality of a generated image.
Only the labeled CANDIDATE IMAGE is provided.
Evaluate whether the image is visually clean, structurally coherent, and free of generation defects.
Use only visible pixel evidence.
Judge the candidate image independently, without assuming any unseen reference, prompt, context image, task, or intended camera pose.\par
\textbf{Evaluate:}
- malformed, melted, stretched, fused, or broken objects and structures;
- internally inconsistent room, object, perspective, or surface geometry;
- duplicated or unnaturally repeated elements;
- seams, corrupted regions, incomplete rendering, and white or black borders;
- unintended text, watermarks, blank regions, or padding;
- unintended blur, pixelation, noise, aliasing, or compression artifacts;
- implausible edges, lighting discontinuities, and inconsistent shadows.\par
\textbf{Do NOT evaluate:}
- whether the image matches an unseen reference, prompt, task, or intended viewpoint;
- whether its objects or layout are correct for an external target;
- semantic fidelity or spatial fidelity;
- aesthetic preference or similarity to a particular rendering style.\par
Do not reward an image merely because it looks attractive, polished, colorful, or cinematic.
Do not penalize unusual content when it is internally coherent.
Judge coherence within the image's own visual domain.\par
\textbf{Visual Quality scoring:}
10 = Technically clean and internally coherent. No meaningful deformation, broken geometry, duplication, border, seam, corruption, or unintended artifact is visible.
8 = High visual quality. Only minor and localized defects are visible, without affecting important objects or structures.
6 = Generally usable, but one prominent defect or several noticeable local artifacts are present.
4 = Multiple prominent defects affect important regions, such as malformed objects, inconsistent geometry, repeated structures, seams, blur, or borders.
2 = Severe and widespread distortion, incomplete rendering, broken perspective, large blank regions, or obvious corruption makes the image difficult to use.
0 = Blank, invalid, unreadable, almost completely corrupted, or a failed generation.\par
\textbf{Visual Quality score constraints:}
- One obvious structural deformation in a salient region prevents a score above 7.
- Multiple visible deformations, repetitions, seams, or broken structures prevent a score above 5.
- A large white or black border, incomplete rendering, or corrupted region prevents a score above 4.
- A score of 9 or 10 requires careful inspection to reveal no meaningful defect.
- Attractive lighting, color, or visual polish alone must not increase the score.\par
Use an integer from 0 to 10. Integers between the stated anchors may be used when justified.
Apply all score constraints before producing the final score.\par
The reason must identify the most important visible defect or, when no meaningful defect exists, explicitly state that the image is visually coherent.
Keep the reason to one short sentence.\par
{\raggedright
Return ONLY valid JSON in exactly this format:
\judgejson{\{ "visual\_quality\_score": \textless{}integer 0-10\textgreater{}, "visual\_quality\_reason": "\textless{}short evidence-based reason\textgreater{}" \}}\par}
% END EXACT PROMPT: build_quality_prompt
\end{tcolorbox}
\endgroup

%% file: sections/mindcue_implementation_details.tex
MindCue uses a target-view hypothesis to identify useful additional evidence
while preserving the complete core.\ %
Following \Cref{sec:mindcue}, $\mathcal I_i$ indexes the ordered input
pool, $\mathcal F_i$ the core, and $\mathcal D_i$ the additional views.\ %
The selector uses only input RGB images, the query, task label, and core and
anchor metadata, with no access to held-out targets, sensor depth,
ground-truth poses, generated images, or evaluation scores.\ %

\subsection{Scene Evidence}
Frozen VGGT jointly predicts camera poses, intrinsics, depth, and confidence,
placing the input views in a shared reconstruction.\ %
For semantics-dominant tasks, a fixed parser extracts queried entities and
frozen SAM~3 segments them.\ %
After confidence filtering and duplicate suppression, masks are lifted into
the reconstruction and grouped into object clusters.\ %
Each cluster must include a core observation, while additional views can
refine its geometry.\ %
Geometry-dominant tasks use the predicted geometry without object masks.\ %

\subsection{Target-View Hypothesis}
\raggedbottom
The query and scene evidence define a hypothesized target camera
$\hat{\mathcal C}^{\star}_i$ with pose and intrinsics.\ %
This camera guides evidence selection; it is not supplied to the generator
as a pose condition.\ %
For tasks with an explicit anchor, MindCue updates the first input camera
according to the requested motion or object reference.\ %
ROG and MCVI instead compare candidates derived from predicted objects and
input-camera positions, as detailed in \Cref{tab:mindcue-camera-solvers}.\ %
World-up is derived from predicted camera poses.\ %

\input{tables/mindcue_virtual_camera_solvers}

\Needspace{5\baselineskip}
ROG ranks candidates by a fixed weighted combination of relation consistency,
target confidence, and proximity to the source input camera.\ %
MCVI instead combines facing-object centering, satisfaction of available
constraints, and observed semantic support.\ %
ROG requires evidence for every parsed relation object, whereas MCVI omits
unsupported auxiliary constraints and treats generic \emph{between} cues as
proximity to both objects.\ %

\subsection{Reliable Target Coverage}
To measure target coverage, we construct sparse per-view point sets
$\hat{\mathcal P}_{iv}$ through depth sampling, confidence filtering,
and unprojection.\ %
Projecting their union through $\hat{\mathcal C}^{\star}_i$ retains the nearest
depth in each occupied grid cell, forming the virtual surface $\mathcal S_i$
in \Cref{eq:mindcue-target-surface}.\ %
View $v$ supports a cell when one of its projected points agrees with this
surface within a depth tolerance.\ %
Thus, support requires depth agreement, not projection overlap alone.\ %

\Needspace{10\baselineskip}
Let $\mathcal V_{iv}\subseteq\mathcal S_i$ be the supported cells and
$\mathcal V_{i,-v}=\bigcup_{u\in\mathcal I_i\setminus\{v\}}\mathcal V_{iu}$
the support from all other input views.\ %
For a nonempty surface, reliable coverage is
\begin{equation}
R_i(v)
=\frac{|\mathcal V_{iv}\cap\mathcal V_{i,-v}|}{|\mathcal S_i|}
 +\min\!\left\{
 \frac{|\mathcal V_{iv}\setminus\mathcal V_{i,-v}|}{|\mathcal S_i|},
 \eta\right\}.
\label{eq:mindcue-coverage}
\end{equation}
The first term credits cross-view support; the second caps the total
contribution of evidence supplied only by the candidate.\ %
We use $\eta=0.05$ and retain every additional view whose reliable coverage
reaches $\tau=0.20$.\ %
Cells are counted equally, and support is computed once from the original
pool, including views subsequently discarded.\ %
Unlike diagnostic Target Coverage in \Cref{sec:context-results},
this score uses only predicted input geometry and a hypothesized camera.\ %

\subsection{Connectivity and Final Selection}
MindCue constructs the retained additional-view set $A_i$ while preserving
the complete core $\mathcal F_i$.\ %
If the virtual surface $\mathcal S_i$ is empty, $A_i=\varnothing$ and the
generator receives only the core.\ %
Otherwise, it first retains every additional view $v\in\mathcal D_i$ with
reliable target coverage $R_i(v)\geq\tau=0.20$.\ %

For tasks with an explicit anchor, one below-threshold view may also be
retained to connect the core to the target hypothesis.\ %
Candidates must meet minimum coverage and connection-score requirements
and pass nonredundancy checks against the core and coverage-selected views.\ %
We select the eligible candidate maximizing
\begin{equation}
Q_i(v)=\sqrt{R_i(v)\max_{u\in\mathcal F_i}\delta_i(v,u)}.
\label{eq:mindcue-connection}
\end{equation}
Here $\delta_i(v,u)$ is the larger of the two directed overlaps: the fraction
of source points reprojecting into the other view with consistent depth,
or zero for an empty source.\ %
The selected view forms $X_i$; without an anchor or an eligible candidate,
$X_i=\varnothing$.\ %

For a nonempty surface, the final additional-view set is
$A_i=\{v\in\mathcal D_i:R_i(v)\geq\tau\}\cup X_i$,
as defined in \Cref{eq:mindcue-selection}.\ %
All coverage-selected additions are retained; connectivity adds at most
one further view.\ %
The generator receives the original query and the pool views indexed by
$\mathcal F_i\cup A_i$, preserving their original order and every core view.\ %

\subsection{Experimental Controls}
Our experiments use $C_{10,i}$ as the pool and $C_{6,i}$ as the core.\ %
RandDrop randomly selects additional views to match MindCue's per-sample
count, whereas CounterSel retains the additions rejected by MindCue and is
not count-matched.\ %
Both preserve the core and input order, with generator checkpoints and
evaluation references fixed across policies.\ %
MindCue uses the same selection for BAGEL and BAGEL-SFT.\ %

\input{sections/mindcue_runtime}

%% file: tables/mindcue_virtual_camera_solvers.tex
% Task-specific camera hypotheses used by MindCue.
\begin{table*}[!t]
\centering
\caption{Task-specific target-view hypotheses in MindCue.}
\label{tab:mindcue-camera-solvers}
\begingroup
\footnotesize
\setlength{\tabcolsep}{4pt}
\renewcommand{\arraystretch}{1.08}
\begin{tabularx}{\textwidth}{@{}>{\raggedright\arraybackslash}p{0.075\textwidth}
>{\raggedright\arraybackslash}X@{}}
\toprule
\textbf{Task} & \textbf{Target-view hypothesis} \\
\midrule
\multicolumn{2}{@{}l}{\textcolor{MindDirect}{\textbf{Geometry-Dominant Tasks}}}\\[2pt]
Trans.
& Move from the reference position along its horizontal camera axes by the
requested displacements, keeping its orientation and intrinsics. \\
Rot.
& Keep the reference position and intrinsics. Rotate about world-up by the
requested yaw, with left positive and right negative. A turn-around is
$180^\circ$. \\
Comp.
& Apply translations and rotations in query order from the reference pose.
Each displacement uses the current horizontal camera axes, accounting for
preceding rotations. Keep the reference intrinsics. \\
\midrule
\multicolumn{2}{@{}l}{\textcolor{MindGrounded}{\textbf{Semantics-Dominant Tasks}}}\\[2pt]
ROG
& For each core-supported target centroid, place cameras at the requested
distance along rays toward input-camera centers. Face the centroid and
inherit source intrinsics; select the highest-scoring object--camera pair. \\
TFR
& Keep the reference center, roll, and intrinsics, and face the
core-supported target centroid. \\
OCR
& Orbit the reference camera around the target centroid about world-up by
the requested angle, preserving distance, roll, and intrinsics; face the target. \\
MCVI
& For each facing-object candidate, reorient cameras where it is detected
toward its centroid, retaining centers and intrinsics; select the highest-scoring pair.
\\
\bottomrule
\end{tabularx}
\endgroup
\end{table*}

%% file: sections/mindcue_runtime.tex
\subsection{Runtime}
\label{sec:mindcue-runtime}
On a single NVIDIA A100-SXM4-80GB, MindCue adds 1.44~s per sample on average,
raising the summed latency from 42.48~s to 43.91~s with identical selected
images and order (\Cref{tab:mindcue-runtime}).\ %
The 3.4\% overhead is the ratio of unrounded weighted mean MindCue and BAGEL
generation times.\ %
VGGT reconstruction and SAM~3 segmentation with mask lifting contribute
0.57~s and 0.82~s on average, respectively.\ %
The SAM~3 mean includes zero time for geometry-dominant tasks.\ %

We use 63 samples stratified by task and dataset, with three repetitions.\ %
Dataset means are weighted by their release frequencies within each task,
then averaged equally across the seven tasks.\ %
Models are preloaded and timed separately with GPU synchronization.\ %
These stage-wise sums exclude RGB reading and preprocessing, model loading and
warmup, inter-stage handoff, output saving, and evaluation.\ %

\input{tables/mindcue_runtime}

Including RGB reading and initial preprocessing in both MindCue and BAGEL
timings gives an additional 1.86~s, or 4.4\%.\ %
These results indicate that MindCue adds only modest overhead relative to
generation under this profiling setup.\ %

%% file: tables/mindcue_runtime.tex
\begin{table}[!htb]
\centering
\caption{Runtime on a single A100 80GB. View counts refer to generator inputs, with identical images and order in the selected-view rows. Means are weighted and rounded independently.}
\label{tab:mindcue-runtime}
\begin{tabular}{lrr}
\toprule
Configuration & Mean views & Mean (s) \\
\midrule
BAGEL, $C_6$ & 6 & 38.528 \\
BAGEL, $C_{10}$ & 10 & 50.625 \\
BAGEL, selected views & 7.30 & 42.476 \\
MindCue + BAGEL, same views & 7.30 & 43.913 \\
\midrule
\multicolumn{2}{l}{MindCue component} & Mean (s) \\
\midrule
\multicolumn{2}{l}{VGGT reconstruction} & 0.572 \\
\multicolumn{2}{l}{SAM~3 segmentation and mask lifting} & 0.819 \\
\multicolumn{2}{l}{Query parsing, camera planning, and selection} & 0.046 \\
\midrule
\multicolumn{2}{l}{\textbf{Total MindCue overhead (+3.4\%)}} & \textbf{1.438} \\
\bottomrule
\end{tabular}
\end{table}

%% file: tables/base_generators_c6_task_component_metrics.tex
\begin{table}[H]
\centering
\caption{Component scores on MindView for Translation and Rotation.}
\label{tab:appendix-baseline-task-metrics}
\label{tab:appendix-sft-task-metrics}
\label{tab:final-sft-c6-taskwise}
\label{tab:baseline-task-components-1}
\begingroup
\fontsize{8}{9}\selectfont
\setlength{\tabcolsep}{1.4pt}
\renewcommand{\arraystretch}{1.02}
\begin{tabular*}{\textwidth}{@{\extracolsep{\fill}}lrrrrrrrrrrrr@{}}
\toprule
\multirow{2}{*}{Model} & \multicolumn{6}{c}{Trans.} & \multicolumn{6}{c}{Rot.} \\
\cmidrule(lr){2-7}\cmidrule(lr){8-13}
 & DINO-I & Sem. & Spa. & Qual. & RScore & TScore & DINO-I & Sem. & Spa. & Qual. & RScore & TScore \\
\midrule
BAGEL & 47.7 & 44.2 & 34.5 & 48.6 & 47.8 & 27.5 & 45.8 & 41.4 & 30.4 & 46.4 & 33.7 & 35.1 \\
FLUX.2 & 53.2 & 49.8 & 40.9 & 85.3 & 59.0 & 29.7 & 46.7 & 42.4 & 33.9 & 78.5 & 31.9 & 48.7 \\
Qwen-Image-Edit-2511\textsuperscript{\ensuremath{\dagger}} & 57.3 & 51.8 & 41.0 & 67.3 & 67.8 & 35.4 & 43.1 & 36.9 & 29.3 & 69.8 & 32.2 & 63.8 \\
Emu3.5\textsuperscript{\ensuremath{\dagger}} & 52.0 & 46.0 & 38.0 & 69.6 & \textbf{80.1} & \underline{48.4} & 38.7 & 32.0 & 24.5 & 63.3 & 30.3 & \underline{72.5} \\
DreamOmni2 & 56.3 & 51.3 & 41.8 & 61.8 & 74.2 & 41.1 & 39.8 & 31.8 & 25.5 & 60.8 & 23.7 & 63.7 \\
SenseNova-U1 & 51.4 & 45.5 & 37.9 & 59.0 & 46.9 & 24.5 & 48.7 & 41.5 & 34.0 & 59.7 & 20.4 & 36.3 \\
SenseNova-U1.5 & 57.8 & 50.7 & 41.8 & 71.3 & 62.9 & 37.2 & 48.9 & 43.4 & 35.2 & 65.7 & 38.0 & 49.4 \\
Nano Banana & 52.4 & 50.5 & 41.0 & 74.1 & 47.8 & 29.9 & 53.3 & 48.4 & 37.3 & 64.5 & 38.2 & 46.8 \\
Nano Banana Pro & 55.5 & \underline{55.9} & 44.4 & 82.4 & 53.0 & 35.9 & \underline{59.3} & \underline{54.9} & \underline{43.5} & 74.6 & 44.8 & 45.1 \\
GPT-Image-1.5 & 52.8 & 52.5 & 42.4 & \underline{85.4} & 52.8 & 26.8 & 52.2 & 50.2 & 39.7 & \underline{81.1} & 34.9 & 29.1 \\
GPT-Image-2 & \underline{59.1} & 55.0 & \underline{45.1} & \textbf{87.1} & 63.6 & 32.1 & 57.1 & 52.9 & 42.1 & \textbf{81.8} & \underline{50.0} & 41.3 \\
\midrule
\textbf{BAGEL-SFT} & \textbf{67.1} & \textbf{70.4} & \textbf{59.0} & 63.9 & \underline{78.4} & \textbf{64.0} & \textbf{72.2} & \textbf{70.3} & \textbf{56.2} & 58.0 & \textbf{69.1} & \textbf{83.5} \\
\bottomrule
\end{tabular*}
\endgroup
\end{table}

\begin{table}[H]
\centering
\caption{Component scores on MindView for Composition and Relational Object Grounding.}
\label{tab:baseline-task-components-2}
\begingroup
\fontsize{8}{9}\selectfont
\setlength{\tabcolsep}{1.4pt}
\renewcommand{\arraystretch}{1.02}
\begin{tabular*}{\textwidth}{@{\extracolsep{\fill}}lrrrrrrrrrrrr@{}}
\toprule
\multirow{2}{*}{Model} & \multicolumn{6}{c}{Comp.} & \multicolumn{6}{c}{ROG} \\
\cmidrule(lr){2-7}\cmidrule(lr){8-13}
 & DINO-I & Sem. & Spa. & Qual. & RScore & TScore & DINO-I & Sem. & Spa. & Qual. & RScore & TScore \\
\midrule
BAGEL & 42.7 & 34.9 & 28.0 & 45.7 & 29.8 & 35.5 & 50.9 & 48.8 & 36.2 & 45.2 & 42.5 & 35.4 \\
FLUX.2 & 43.6 & 40.0 & 33.5 & 80.8 & 32.9 & 33.6 & 54.1 & 52.0 & 40.5 & 81.3 & 33.2 & 42.2 \\
Qwen-Image-Edit-2511\textsuperscript{\ensuremath{\dagger}} & 43.7 & 38.3 & 30.7 & 66.5 & 31.6 & 39.0 & 55.2 & 54.6 & 42.0 & 69.3 & 38.4 & 39.8 \\
Emu3.5\textsuperscript{\ensuremath{\dagger}} & 38.0 & 31.3 & 26.9 & 63.8 & 26.8 & \underline{57.3} & 49.0 & 45.5 & 32.8 & 58.5 & 29.9 & 39.1 \\
DreamOmni2 & 36.0 & 28.5 & 24.4 & 64.0 & 19.5 & 42.6 & 35.4 & 29.4 & 23.6 & 70.5 & 11.2 & 32.3 \\
SenseNova-U1 & 44.7 & 37.6 & 32.5 & 57.5 & 24.8 & 30.2 & 61.1 & 55.7 & 42.5 & 66.5 & 30.3 & 40.1 \\
SenseNova-U1.5 & 49.4 & 40.2 & 34.1 & 68.2 & 35.4 & 31.9 & 62.3 & 59.2 & 47.7 & 69.3 & 50.9 & \underline{51.0} \\
Nano Banana & 47.1 & 44.9 & 36.5 & 67.5 & 39.3 & 39.5 & 60.5 & 59.4 & 43.8 & 64.6 & 45.3 & 46.9 \\
Nano Banana Pro & 46.8 & \underline{46.1} & 37.6 & 79.0 & 39.5 & 34.9 & 65.6 & 61.1 & 48.8 & \underline{85.2} & 35.7 & 41.5 \\
GPT-Image-1.5 & 47.1 & 45.1 & 37.3 & \underline{82.3} & 40.4 & 25.5 & 66.9 & \underline{66.9} & \underline{53.1} & 82.8 & 45.7 & 39.6 \\
GPT-Image-2 & \underline{49.7} & \underline{46.1} & \underline{38.1} & \textbf{85.2} & \underline{44.6} & 25.3 & \underline{67.1} & 63.6 & 50.2 & \textbf{85.5} & \textbf{55.7} & 34.4 \\
\midrule
\textbf{BAGEL-SFT} & \textbf{57.4} & \textbf{54.3} & \textbf{44.8} & 60.2 & \textbf{56.3} & \textbf{64.8} & \textbf{68.5} & \textbf{71.2} & \textbf{57.3} & 60.9 & \underline{54.6} & \textbf{63.4} \\
\bottomrule
\end{tabular*}
\endgroup
\end{table}

\begin{table}[H]
\centering
\caption{Component scores on MindView for Target-Facing Reorientation and Object-Centric Repositioning.}
\label{tab:baseline-task-components-3}
\begingroup
\fontsize{8}{9}\selectfont
\setlength{\tabcolsep}{1.4pt}
\renewcommand{\arraystretch}{1.02}
\begin{tabular*}{\textwidth}{@{\extracolsep{\fill}}lrrrrrrrrrrrr@{}}
\toprule
\multirow{2}{*}{Model} & \multicolumn{6}{c}{TFR} & \multicolumn{6}{c}{OCR} \\
\cmidrule(lr){2-7}\cmidrule(lr){8-13}
 & DINO-I & Sem. & Spa. & Qual. & RScore & TScore & DINO-I & Sem. & Spa. & Qual. & RScore & TScore \\
\midrule
BAGEL & 43.4 & 42.0 & 30.9 & 45.0 & 32.4 & 38.3 & 61.7 & 59.2 & 44.1 & 50.8 & 27.0 & 22.0 \\
FLUX.2 & 50.5 & 48.5 & 38.3 & 76.1 & 32.7 & 53.4 & 70.0 & 68.4 & 52.9 & 77.6 & 28.4 & 25.8 \\
Qwen-Image-Edit-2511\textsuperscript{\ensuremath{\dagger}} & 48.8 & 45.6 & 35.3 & 75.6 & 31.0 & 43.5 & 74.2 & \underline{79.4} & 60.6 & 61.9 & 31.1 & 34.8 \\
Emu3.5\textsuperscript{\ensuremath{\dagger}} & 40.6 & 40.8 & 29.8 & 73.3 & 25.7 & \underline{59.4} & 62.7 & 61.2 & 43.8 & 68.5 & \underline{37.7} & \underline{40.0} \\
DreamOmni2 & 31.0 & 25.6 & 19.2 & 67.7 & 13.2 & 53.4 & 62.7 & 60.2 & 45.6 & 63.4 & 23.8 & 31.5 \\
SenseNova-U1 & 56.7 & 54.4 & 40.5 & 67.0 & 18.8 & 43.3 & 60.9 & 57.2 & 42.7 & 68.5 & 15.7 & 22.5 \\
SenseNova-U1.5 & 59.3 & 55.2 & 43.5 & 70.3 & 33.9 & 58.3 & 70.2 & 70.0 & 53.6 & 70.5 & 32.5 & 31.3 \\
Nano Banana & 64.7 & 64.1 & 48.2 & 60.2 & 43.3 & 53.6 & 62.5 & 59.4 & 45.7 & 65.7 & 26.1 & 30.3 \\
Nano Banana Pro & 67.2 & 64.5 & 51.2 & 76.2 & 41.1 & 51.3 & 68.3 & 67.7 & 51.9 & 76.9 & 32.2 & 30.9 \\
GPT-Image-1.5 & 66.8 & 68.0 & 52.9 & \underline{79.4} & 42.4 & 46.8 & 67.9 & 72.6 & 54.0 & \underline{83.2} & 23.1 & 24.7 \\
GPT-Image-2 & \textbf{71.5} & \underline{74.5} & \underline{58.9} & \textbf{81.4} & \underline{55.6} & 58.7 & \underline{74.6} & 78.2 & \underline{61.2} & \textbf{85.2} & 37.5 & 37.8 \\
\midrule
\textbf{BAGEL-SFT} & \underline{70.5} & \textbf{80.3} & \textbf{63.3} & 60.3 & \textbf{59.2} & \textbf{65.3} & \textbf{78.9} & \textbf{86.7} & \textbf{69.1} & 63.7 & \textbf{70.0} & \textbf{72.2} \\
\bottomrule
\end{tabular*}
\endgroup
\end{table}

\begin{table}[H]
\centering
\caption{Component scores on MindView for Multi-Cue Viewpoint Inference.}
\label{tab:baseline-task-components-4}
\begingroup
\fontsize{8}{9}\selectfont
\setlength{\tabcolsep}{1.4pt}
\renewcommand{\arraystretch}{1.02}
\begin{tabular*}{\textwidth}{@{\extracolsep{\fill}}lrrrrrr@{}}
\toprule
\multirow{2}{*}{Model} & \multicolumn{6}{c}{MCVI} \\
\cmidrule(lr){2-7}
 & DINO-I & Sem. & Spa. & Qual. & RScore & TScore \\
\midrule
BAGEL & 45.8 & 41.2 & 31.4 & 47.9 & 30.0 & 29.3 \\
FLUX.2 & 47.5 & 45.8 & 36.0 & 82.9 & 30.4 & 36.7 \\
Qwen-Image-Edit-2511\textsuperscript{\ensuremath{\dagger}} & 44.6 & 37.9 & 31.4 & 76.2 & 28.8 & 25.5 \\
Emu3.5\textsuperscript{\ensuremath{\dagger}} & 32.9 & 30.6 & 22.6 & 73.0 & 21.4 & 32.3 \\
DreamOmni2 & 31.0 & 24.9 & 21.2 & 70.1 & 11.3 & 34.5 \\
SenseNova-U1 & 51.7 & 45.8 & 37.8 & 66.3 & 22.5 & 35.6 \\
SenseNova-U1.5 & 58.1 & 51.2 & 42.1 & 72.3 & 38.4 & 43.3 \\
Nano Banana & 55.0 & 51.4 & 39.6 & 67.2 & 33.0 & 37.7 \\
Nano Banana Pro & 57.1 & 50.2 & 40.5 & \underline{85.7} & 25.4 & 33.5 \\
GPT-Image-1.5 & 58.2 & 56.7 & 44.9 & 81.2 & 39.7 & 41.8 \\
GPT-Image-2 & \textbf{61.5} & \underline{60.1} & \underline{47.7} & \textbf{86.2} & \underline{42.6} & \underline{44.0} \\
\midrule
\textbf{BAGEL-SFT} & \underline{61.0} & \textbf{63.8} & \textbf{50.2} & 65.8 & \textbf{44.0} & \textbf{52.5} \\
\bottomrule
\end{tabular*}
\endgroup
\end{table}

%% file: tables/base_generators_context_component_metrics.tex
\begin{table}[H]
\centering
\caption{Task-macro component scores under reduced and expanded contexts.}
\label{tab:appendix-context-metrics}
\label{tab:final-all-contexts}
\label{tab:appendix-sft-context-metrics}
\label{tab:final-sft-all-contexts}
\begingroup
\fontsize{8}{9}\selectfont
\setlength{\tabcolsep}{1.4pt}
\renewcommand{\arraystretch}{1.02}
\begin{tabular*}{\textwidth}{@{\extracolsep{\fill}}lrrrrrrrrrrrr@{}}
\toprule
\multirow{2}{*}{Model} & \multicolumn{6}{c}{$C_3$} & \multicolumn{6}{c}{$C_{10}$} \\
\cmidrule(lr){2-7}\cmidrule(lr){8-13}
 & DINO-I & Sem. & Spa. & Qual. & RScore & TScore & DINO-I & Sem. & Spa. & Qual. & RScore & TScore \\
\midrule
BAGEL & 46.7 & 43.2 & 32.4 & 47.3 & 33.0 & 34.0 & 45.9 & 40.8 & 31.0 & 48.5 & 32.4 & 35.5 \\
FLUX.2 & 49.6 & 46.4 & 37.7 & \textbf{85.9} & 33.7 & 41.5 & 51.5 & 48.7 & 38.1 & 69.3 & 32.5 & 38.0 \\
Qwen-Image-Edit-2511\textsuperscript{\ensuremath{\dagger}} & 50.2 & 47.1 & 37.2 & 69.8 & 33.8 & 38.1 & 51.6 & 48.1 & 37.9 & 71.1 & 36.2 & 38.3 \\
Emu3.5\textsuperscript{\ensuremath{\dagger}} & 44.5 & 38.9 & 28.6 & 59.5 & 34.8 & \underline{46.8} & 44.9 & 39.9 & 30.6 & 68.3 & 33.3 & 49.1 \\
DreamOmni2 & 42.7 & 37.6 & 30.8 & 76.9 & 25.1 & 36.1 & 32.2 & 27.7 & 22.1 & 59.0 & 25.9 & \underline{51.8} \\
SenseNova-U1 & 50.3 & 46.1 & 36.8 & 60.1 & 27.4 & 35.1 & 51.6 & 45.3 & 36.5 & 66.4 & 22.3 & 32.2 \\
SenseNova-U1.5 & 54.8 & 49.1 & 39.5 & 69.8 & 38.8 & 45.2 & 57.3 & 51.8 & 41.6 & 72.1 & 40.6 & 43.0 \\
Nano Banana & 55.1 & 51.6 & 39.7 & 62.8 & 38.4 & 44.2 & 54.5 & 51.1 & 39.7 & 68.9 & 37.1 & 39.3 \\
Nano Banana Pro & 57.2 & 54.3 & 43.4 & 79.3 & 37.0 & 40.9 & 58.2 & 54.9 & 43.5 & 80.8 & 36.6 & 38.6 \\
GPT-Image-1.5 & 57.6 & 55.9 & 44.5 & 82.1 & 38.0 & 37.7 & 57.2 & 56.7 & 44.6 & \underline{83.2} & 35.5 & 31.9 \\
GPT-Image-2 & \underline{59.0} & \underline{57.9} & \underline{46.3} & \underline{84.8} & \underline{45.2} & 41.6 & \underline{61.2} & \underline{59.8} & \underline{47.9} & \textbf{85.1} & \underline{48.5} & 37.5 \\
\midrule
\textbf{BAGEL-SFT} & \textbf{63.7} & \textbf{62.5} & \textbf{50.4} & 61.4 & \textbf{59.4} & \textbf{65.8} & \textbf{67.4} & \textbf{69.5} & \textbf{55.8} & 59.6 & \textbf{61.7} & \textbf{64.9} \\
\bottomrule
\end{tabular*}
\endgroup
\end{table}

%% file: tables/training_free_context_selection_components.tex
\begin{table}[H]
\centering
\caption{Task-macro component scores for the five context policies in \Cref{tab:training_free_effectiveness}.}
\label{tab:selection-component-metrics}
\begingroup
\fontsize{8}{9}\selectfont
\setlength{\tabcolsep}{1.4pt}
\renewcommand{\arraystretch}{1.02}
\begin{tabular*}{\textwidth}{@{\extracolsep{\fill}}llrrrrrr@{}}
\toprule
Model & Policy & DINO-I & Sem. & Spa. & Qual. & RScore & TScore \\
\midrule
\multirow{5}{*}{BAGEL} & Fixed $C_6$ & \underline{48.3} & \underline{44.5} & \underline{33.6} & 47.1 & \underline{34.7} & 31.9 \\
 & Fixed $C_{10}$ & 45.9 & 40.8 & 31.0 & \textbf{48.5} & 32.4 & \textbf{35.5} \\
 & CounterSel & 43.7 & 38.4 & 29.3 & \underline{48.4} & 29.5 & 33.0 \\
 & RandDrop & 47.3 & 42.8 & 32.0 & 47.9 & 34.0 & 33.3 \\
 & \textbf{MindCue} & \textbf{49.2} & \textbf{44.9} & \textbf{33.9} & 47.6 & \textbf{36.6} & \underline{34.7} \\
\midrule
\multirow{5}{*}{BAGEL-SFT} & Fixed $C_6$ & \underline{68.0} & \underline{71.0} & \underline{57.1} & \textbf{61.8} & 61.7 & \underline{66.5} \\
 & Fixed $C_{10}$ & 67.4 & 69.5 & 55.8 & 59.6 & 61.7 & 64.9 \\
 & CounterSel & 66.2 & 67.7 & 54.5 & 60.2 & 61.8 & 65.4 \\
 & RandDrop & 67.8 & 70.4 & 56.6 & 61.3 & \underline{62.0} & 66.2 \\
 & \textbf{MindCue} & \textbf{68.9} & \textbf{71.8} & \textbf{57.7} & \underline{61.5} & \textbf{63.6} & \textbf{67.1} \\
\bottomrule
\end{tabular*}
\endgroup
\end{table}

%% file: sections/limitations.tex
\section{Limitations}
\label{sec:limitations}

MindView uses one real target image per query to maintain a simple, uniform
evaluation protocol across all seven tasks.\ %
This is a trade-off for ROG and MCVI, whose semantic constraints may admit
nearby valid viewpoints.\ %
We reduce ambiguity through object-reference and visibility checks, bounded
view clustering, checks for competing viewpoints, and LLM-assisted and
independent expert review (\Cref{sec:construction-details}).\ %
For example, the recorded matching sets of the 243 released MCVI samples
have median horizontal-position and yaw diameters of only $1.7$\,cm and
$0.93^\circ$.\ %
Feature similarity, graded VLM ratings, and linearly decreasing geometric
scores allow partial credit for modest deviations
(\Cref{sec:metric-details}), so valid views with similar content and
layout within the same compact cluster are expected to receive similar
scores.\ %
Taken together, these ambiguity controls and graded metrics support using a
single representative GT as a reasonable practical compromise for consistent
evaluation across tasks.\ %

%% file: main.bbl
\begin{thebibliography}{42}
\providecommand{\natexlab}[1]{#1}
\providecommand{\url}[1]{\texttt{#1}}
\expandafter\ifx\csname urlstyle\endcsname\relax
  \providecommand{\doi}[1]{doi: #1}\else
  \providecommand{\doi}{doi: \begingroup \urlstyle{rm}\Url}\fi

\bibitem[Baruch et~al.(2021)Baruch, Chen, Dehghan, Dimry, Feigin, Fu, Gebauer, Joffe, Kurz, Schwartz, and Shulman]{arkitscenes2021}
Gilad Baruch, Zhuoyuan Chen, Afshin Dehghan, Tal Dimry, Yuri Feigin, Peter Fu, Thomas Gebauer, Brandon Joffe, Daniel Kurz, Arik Schwartz, and Elad Shulman.
\newblock Arkitscenes: A diverse real-world dataset for 3d indoor scene understanding using mobile rgb-d data.
\newblock In \emph{Thirty-fifth Conference on Neural Information Processing Systems Datasets and Benchmarks Track}, 2021.

\bibitem[{Black Forest Labs}(2026)]{exp_flux2_klein2026}
{Black Forest Labs}.
\newblock {FLUX.2 [klein]}: Towards interactive visual intelligence.
\newblock \url{https://bfl.ai/blog/flux2-klein-towards-interactive-visual-intelligence}, January 2026.
\newblock Evaluated checkpoint: \url{https://huggingface.co/black-forest-labs/FLUX.2-klein-9B}. Accessed 2026-09-19.

\bibitem[Cai et~al.(2026)Cai, Wang, Gu, Pu, Xu, Wang, Yin, Yang, Wei, Zhou, Sun, Pang, Li, Qian, Lin, Shi, Deng, Han, Chen, Fan, Deng, Lu, Pan, Li, Liu, Wang, Lin, and Yang]{exp_sensenova_si2026}
Zhongang Cai, Ruisi Wang, Chenyang Gu, Fanyi Pu, Junxiang Xu, Yubo Wang, Wanqi Yin, Zhitao Yang, Chen Wei, Tongxi Zhou, Qingping Sun, Hui~En Pang, Jiaqi Li, Oscar Qian, Zhiqian Lin, Xuanke Shi, Kewang Deng, Xiaoyang Han, Zukai Chen, Xiangyu Fan, Hanming Deng, Lewei Lu, Liang Pan, Bo~Li, Ziwei Liu, Quan Wang, Dahua Lin, and Lei Yang.
\newblock Scaling spatial intelligence with multimodal foundation models.
\newblock In \emph{Proceedings of the IEEE/CVF Conference on Computer Vision and Pattern Recognition (CVPR)}, pp.\  7879--7890, June 2026.

\bibitem[Carion et~al.(2026)Carion, Gustafson, Hu, Debnath, Hu, Suris Coll-Vinent, Ryali, Alwala, Khedr, Huang, Lei, Ma, Guo, Kalla, Marks, Greer, Wang, Sun, R{\"a}dle, Afouras, Mavroudi, Xu, Wu, Zhou, Momeni, Hazra, Ding, Vaze, Porcher, Li, Li, Kamath, Cheng, Doll{\'a}r, Ravi, Saenko, Zhang, and Feichtenhofer]{sam32026}
Nicolas Carion, Laura Gustafson, Yuan-Ting Hu, Shoubhik Debnath, Ronghang Hu, Didac Suris Coll-Vinent, Chaitanya Ryali, Kalyan~Vasudev Alwala, Haitham Khedr, Andrew Huang, Jie Lei, Tengyu Ma, Baishan Guo, Arpit Kalla, Markus Marks, Joseph Greer, Meng Wang, Peize Sun, Roman R{\"a}dle, Triantafyllos Afouras, Effrosyni Mavroudi, Katherine Xu, Tsung-Han Wu, Yu~Zhou, Liliane Momeni, Rishi Hazra, Shuangrui Ding, Sagar Vaze, Francois Porcher, Feng Li, Siyuan Li, Aishwarya Kamath, Ho~Kei Cheng, Piotr Doll{\'a}r, Nikhila Ravi, Kate Saenko, Pengchuan Zhang, and Christoph Feichtenhofer.
\newblock {SAM 3}: Segment anything with concepts.
\newblock In \emph{International Conference on Learning Representations}, 2026.
\newblock URL \url{https://proceedings.iclr.cc/paper_files/paper/2026/file/e0982cbc81401df3430ee1ff780dc7a2-Paper-Conference.pdf}.

\bibitem[Chen et~al.(2025)Chen, Chen, Wu, Wang, Lang, Sushko, Jiang, Wan, and Krishna]{multiref2025}
Ruoxi Chen, Dongping Chen, Siyuan Wu, Sinan Wang, Shiyun Lang, Petr Sushko, Gaoyang Jiang, Yao Wan, and Ranjay Krishna.
\newblock Multiref: Controllable image generation with multiple visual references.
\newblock \emph{arXiv preprint arXiv:2508.06905}, 2025.

\bibitem[Cui et~al.(2025)Cui, Chen, Deng, Huang, Li, Liu, Liu, Luo, Wang, Wang, Wang, Wang, Zhang, Zhao, Pan, Li, Hao, Ma, Chen, Ao, Huang, Wang, and Wang]{exp_emu35_2025}
Yufeng Cui, Honghao Chen, Haoge Deng, Xu~Huang, Xinghang Li, Jirong Liu, Yang Liu, Zhuoyan Luo, Jinsheng Wang, Wenxuan Wang, Yueze Wang, Chengyuan Wang, Fan Zhang, Yingli Zhao, Ting Pan, Xianduo Li, Zecheng Hao, Wenxuan Ma, Zhuo Chen, Yulong Ao, Tiejun Huang, Zhongyuan Wang, and Xinlong Wang.
\newblock {Emu3.5}: Native multimodal models are world learners.
\newblock \emph{arXiv preprint arXiv:2510.26583}, 2025.

\bibitem[Dai et~al.(2017)Dai, Chang, Savva, Halber, Funkhouser, and Nie{\ss}ner]{scannet2017}
Angela Dai, Angel~X. Chang, Manolis Savva, Maciej Halber, Thomas Funkhouser, and Matthias Nie{\ss}ner.
\newblock Scannet: Richly-annotated 3d reconstructions of indoor scenes.
\newblock In \emph{Proceedings of the IEEE Conference on Computer Vision and Pattern Recognition}, 2017.

\bibitem[Deng et~al.(2025)Deng, Zhu, Li, Gou, Li, Wang, Zhong, Yu, Nie, Song, Shi, and Fan]{exp_bagel2025}
Chaorui Deng, Deyao Zhu, Kunchang Li, Chenhui Gou, Feng Li, Zeyu Wang, Shu Zhong, Weihao Yu, Xiaonan Nie, Ziang Song, Guang Shi, and Haoqi Fan.
\newblock Emerging properties in unified multimodal pretraining.
\newblock \emph{arXiv preprint arXiv:2505.14683}, 2025.

\bibitem[Diao et~al.(2026{\natexlab{a}})Diao, Wang, Ding, Deng, Chen, Zhang, Wang, Tong, Fan, Wang, et~al.]{exp_sensenova_u15_2026}
Haiwen Diao, Jiahao Wang, Chenjing Ding, Hanming Deng, Jiangnan Chen, Ruixi Zhang, Ruohui Wang, Wenwen Tong, Xiangyu Fan, Yubo Wang, et~al.
\newblock {SenseNova-U1.5}: Towards native unified visual intelligence.
\newblock \emph{arXiv preprint arXiv:2609.11929}, 2026{\natexlab{a}}.

\bibitem[Diao et~al.(2026{\natexlab{b}})Diao, Wu, Deng, Wang, Bai, Wu, Fan, Ye, Tong, Fan, et~al.]{exp_sensenova_u1_2026}
Haiwen Diao, Penghao Wu, Hanming Deng, Jiahao Wang, Shihao Bai, Silei Wu, Weichen Fan, Wenjie Ye, Wenwen Tong, Xiangyu Fan, et~al.
\newblock {SenseNova-U1}: Unifying multimodal understanding and generation with {NEO-unify} architecture.
\newblock \emph{arXiv preprint arXiv:2605.12500}, 2026{\natexlab{b}}.

\bibitem[Fortin et~al.(2025)Fortin, Vernade, Kampf, and Reshi]{exp_nanobanana2025}
Alisa Fortin, Guillaume Vernade, Kat Kampf, and Ammaar Reshi.
\newblock Introducing {Gemini 2.5 Flash Image}, our state-of-the-art image model.
\newblock Google Developers Blog, August 2025.
\newblock URL \url{https://developers.googleblog.com/introducing-gemini-2-5-flash-image/}.
\newblock Official announcement of Nano Banana. Published August 26, 2025. Accessed 2026-09-19.

\bibitem[Jin et~al.(2025)Jin, Jiang, Tan, Zhang, Bi, Zhang, Luan, Snavely, and Xu]{lvsm2025}
Haian Jin, Hanwen Jiang, Hao Tan, Kai Zhang, Sai Bi, Tianyuan Zhang, Fujun Luan, Noah Snavely, and Zexiang Xu.
\newblock Lvsm: A large view synthesis model with minimal 3d inductive bias.
\newblock \emph{arXiv preprint arXiv:2410.17242}, 2025.

\bibitem[Kong et~al.(2024)Kong, Liu, Lyu, Taher, Qi, and Davison]{eschernet2024}
Xin Kong, Shikun Liu, Xiaoyang Lyu, Marwan Taher, Xiaojuan Qi, and Andrew~J. Davison.
\newblock Eschernet: A generative model for scalable view synthesis.
\newblock In \emph{Proceedings of the IEEE/CVF Conference on Computer Vision and Pattern Recognition}, 2024.

\bibitem[Li et~al.(2026)Li, Ikehata, Kurita, and Sato]{capframe2026}
Jirong Li, Satoshi Ikehata, Shuhei Kurita, and Ikuro Sato.
\newblock {CapFrame: Text-Instructed Viewpoint Grounding in 3D Gaussian Scenes via Geometric Pseudo Labels}.
\newblock \emph{arXiv preprint arXiv:2608.30342}, 2026.

\bibitem[Li et~al.(2023)Li, Cao, Wang, Qi, Cheng, and Shan]{photomaker2023}
Zhen Li, Mingdeng Cao, Xintao Wang, Zhongang Qi, Ming-Ming Cheng, and Ying Shan.
\newblock Photomaker: Customizing realistic human photos via stacked id embedding.
\newblock \emph{arXiv preprint arXiv:2312.04461}, 2023.

\bibitem[Loshchilov \& Hutter(2019)Loshchilov and Hutter]{adamw2019}
Ilya Loshchilov and Frank Hutter.
\newblock Decoupled weight decay regularization.
\newblock In \emph{International Conference on Learning Representations}, 2019.
\newblock URL \url{https://openreview.net/forum?id=Bkg6RiCqY7}.

\bibitem[Luo et~al.(2026)Luo, Zheng, Shen, Xin, Xu, Liu, Zhang, Zhou, Xie, Li, Cao, Pu, He, Fu, Liu, Qiao, Zhai, Cao, and Liu]{micebench2026}
Siqi Luo, Huayu Zheng, Jianghan Shen, Yi~Xin, Luxin Xu, Jiyao Liu, Xinyu Zhang, Hang Zhou, Pengyu Xie, Xiaohui Li, Shuo Cao, Yuandong Pu, Junjun He, Bin Fu, Yihao Liu, Yu~Qiao, Guangtao Zhai, Yuewen Cao, and Xiaohong Liu.
\newblock {MICE-Bench}: A challenging and comprehensive benchmark for multi-reference image creation and editing.
\newblock In \emph{International Conference on Machine Learning}, 2026.

\bibitem[Mildenhall et~al.(2020)Mildenhall, Srinivasan, Tancik, Barron, Ramamoorthi, and Ng]{nerf2020}
Ben Mildenhall, Pratul~P. Srinivasan, Matthew Tancik, Jonathan~T. Barron, Ravi Ramamoorthi, and Ren Ng.
\newblock Nerf: Representing scenes as neural radiance fields for view synthesis.
\newblock In \emph{European Conference on Computer Vision}, 2020.

\bibitem[{OpenAI}(2025)]{exp_gpt_image15_2025}
{OpenAI}.
\newblock {GPT-Image-1.5}.
\newblock \url{https://developers.openai.com/api/docs/models/gpt-image-1.5}, 2025.
\newblock Official model documentation. Accessed 2026-09-19.

\bibitem[{OpenAI}(2026{\natexlab{a}})]{exp_gpt_image2_2026}
{OpenAI}.
\newblock {GPT-Image-2}.
\newblock \url{https://developers.openai.com/api/docs/models/gpt-image-2}, 2026{\natexlab{a}}.
\newblock Official model documentation. Accessed 2026-09-19.

\bibitem[{OpenAI}(2026{\natexlab{b}})]{gpt562026}
{OpenAI}.
\newblock {GPT-5.6 Sol}.
\newblock \url{https://developers.openai.com/api/docs/models/gpt-5.6-sol}, 2026{\natexlab{b}}.
\newblock Official model documentation. Accessed 2026-09-24.

\bibitem[Oquab et~al.(2023)Oquab, Darcet, Moutakanni, Vo, Szafraniec, Khalidov, Fernandez, Haziza, Massa, El-Nouby, et~al.]{dinov22023}
Maxime Oquab, Timothee Darcet, Theo Moutakanni, Huy~V. Vo, Marc Szafraniec, Vasil Khalidov, Pierre Fernandez, Daniel Haziza, Francisco Massa, Alaaeldin El-Nouby, et~al.
\newblock Dinov2: Learning robust visual features without supervision.
\newblock \emph{arXiv preprint arXiv:2304.07193}, 2023.

\bibitem[Oshima et~al.(2026)Oshima, Miyake, Matsutani, Iwasawa, Suzuki, Matsuo, and Furuta]{multibanana2026}
Yuta Oshima, Daiki Miyake, Kohsei Matsutani, Yusuke Iwasawa, Masahiro Suzuki, Yutaka Matsuo, and Hiroki Furuta.
\newblock {MultiBanana}: A challenging benchmark for multi-reference text-to-image generation.
\newblock In \emph{Proceedings of the IEEE/CVF Conference on Computer Vision and Pattern Recognition}, 2026.

\bibitem[{Qwen Team}(2025{\natexlab{a}})]{exp_qwen_image_edit2511}
{Qwen Team}.
\newblock {Qwen-Image-Edit-2511} model card.
\newblock \url{https://huggingface.co/Qwen/Qwen-Image-Edit-2511}, 2025{\natexlab{a}}.
\newblock Accessed 2026-09-19.

\bibitem[{Qwen Team}(2025{\natexlab{b}})]{qwen3vl_judge_model}
{Qwen Team}.
\newblock {Qwen3-VL-30B-A3B-Instruct} model card.
\newblock \url{https://huggingface.co/Qwen/Qwen3-VL-30B-A3B-Instruct}, 2025{\natexlab{b}}.
\newblock Accessed 2026-09-16.

\bibitem[Raisinghani(2025)]{exp_nanobananapro2025}
Naina Raisinghani.
\newblock Introducing {Nano Banana Pro}.
\newblock Google, The Keyword, November 2025.
\newblock URL \url{https://blog.google/innovation-and-ai/products/nano-banana-pro/}.
\newblock Official announcement of Gemini 3 Pro Image. Published November 20, 2025. Accessed 2026-09-19.

\bibitem[Ray et~al.(2025)Ray, Duan, Brown, Tan, Bashkirova, Hendrix, Ehsani, Kembhavi, Plummer, Krishna, Zeng, and Saenko]{exp_sat2025}
Arijit Ray, Jiafei Duan, Ellis~L. Brown, II, Reuben Tan, Dina Bashkirova, Rose Hendrix, Kiana Ehsani, Aniruddha Kembhavi, Bryan~A. Plummer, Ranjay Krishna, Kuo-Hao Zeng, and Kate Saenko.
\newblock {SAT}: Dynamic spatial aptitude training for multimodal language models.
\newblock In \emph{Second Conference on Language Modeling}, 2025.
\newblock URL \url{https://openreview.net/forum?id=DW8U8ZWa1U}.

\bibitem[Santos et~al.(1997)Santos, Ortiz~de Sol{\'o}rzano, Vaquero, Pe{\~n}a, Malpica, and del Pozo]{santos1997autofocus}
A.~Santos, C.~Ortiz~de Sol{\'o}rzano, J.~J. Vaquero, J.~M. Pe{\~n}a, N.~Malpica, and F.~del Pozo.
\newblock Evaluation of autofocus functions in molecular cytogenetic analysis.
\newblock \emph{Journal of Microscopy}, 188\penalty0 (3):\penalty0 264--272, 1997.
\newblock \doi{10.1046/j.1365-2818.1997.2630819.x}.

\bibitem[Sargent et~al.(2024)Sargent, Li, Shah, Herrmann, Yu, Zhang, Chan, Lagun, Fei-Fei, Sun, and Wu]{zeronvs2024}
Kyle Sargent, Zizhang Li, Tanmay Shah, Charles Herrmann, Hong-Xing Yu, Yunzhi Zhang, Eric~Ryan Chan, Dmitry Lagun, Li~Fei-Fei, Deqing Sun, and Jiajun Wu.
\newblock Zeronvs: Zero-shot 360-degree view synthesis from a single image.
\newblock \emph{arXiv preprint arXiv:2310.17994}, 2024.

\bibitem[Tan et~al.(2026)Tan, Wang, Zhou, Quan, Li, Chen, Zheng, Wang, and Dong]{objectuni2026}
Mining Tan, Yinuo Wang, Ziqi Zhou, Weize Quan, Sifei Li, Jingdong Chen, DanDan Zheng, Libin Wang, and Weiming Dong.
\newblock {Object-Uni: A Unified Model for Object-Centric Spatial Understanding and Controllable Generation}.
\newblock \emph{arXiv preprint arXiv:2608.22757}, 2026.

\bibitem[Wang et~al.(2025{\natexlab{a}})Wang, Chen, Karaev, Vedaldi, Rupprecht, and Novotny]{vggt2025}
Jianyuan Wang, Minghao Chen, Nikita Karaev, Andrea Vedaldi, Christian Rupprecht, and David Novotny.
\newblock Vggt: Visual geometry grounded transformer.
\newblock \emph{arXiv preprint arXiv:2503.11651}, 2025{\natexlab{a}}.

\bibitem[Wang et~al.(2025{\natexlab{b}})Wang, Yin, Zhang, Zhang, Wang, Wang, Zhang, Chandrasegaran, Liu, Krishna, Xie, Wu, Fei-Fei, and Li]{exp_mindcube2025}
Qineng Wang, Baiqiao Yin, Pingyue Zhang, Jianshu Zhang, Kangrui Wang, Zihan Wang, Jieyu Zhang, Keshigeyan Chandrasegaran, Han Liu, Ranjay Krishna, Saining Xie, Jiajun Wu, Li~Fei-Fei, and Manling Li.
\newblock {MindCube}: Spatial mental modeling from limited views.
\newblock \emph{arXiv preprint arXiv:2506.21458}, 2025{\natexlab{b}}.

\bibitem[Wang et~al.(2025{\natexlab{c}})Wang, Xu, Zhang, Pang, Du, Zhao, and Zhao]{genspace2025}
Zehan Wang, Jiayang Xu, Ziang Zhang, Tianyu Pang, Chao Du, Hengshuang Zhao, and Zhou Zhao.
\newblock Genspace: Benchmarking spatially-aware image generation.
\newblock \emph{arXiv preprint arXiv:2505.24870}, 2025{\natexlab{c}}.

\bibitem[Wen et~al.(2026)Wen, Li, Zhang, Lei, Chen, Fan, Zhang, Wang, Qiu, Li, Liu, Shan, Yang, and Shen]{unig2u2026}
Zimo Wen, Boxiu Li, Wanbo Zhang, Junxiang Lei, Xiaoyu Chen, Yijia Fan, Qi~Zhang, Yujiang Wang, Lili Qiu, Bo~Li, Ziwei Liu, Caihua Shan, Yifan Yang, and Yifei Shen.
\newblock {UniG2U-Bench}: Do unified models advance multimodal understanding?
\newblock \emph{arXiv preprint arXiv:2603.03241}, 2026.

\bibitem[Wu et~al.(2025)Wu, Li, Zhou, Lin, Gao, Yan, Yin, Bai, Xu, Chen, Chen, Tang, Zhang, Wang, Yang, Yu, Cheng, Liu, Li, Zhang, Meng, Wei, Ni, Chen, Cao, Peng, Qu, Wu, Wang, Yu, Wen, Feng, Xu, Wang, Zhang, Zhu, Wu, Cai, and Liu]{exp_qwen_image2025}
Chenfei Wu, Jiahao Li, Jingren Zhou, Junyang Lin, Kaiyuan Gao, Kun Yan, Sheng-ming Yin, Shuai Bai, Xiao Xu, Yilei Chen, Yuxiang Chen, Zecheng Tang, Zekai Zhang, Zhengyi Wang, An~Yang, Bowen Yu, Chen Cheng, Dayiheng Liu, Deqing Li, Hang Zhang, Hao Meng, Hu~Wei, Jingyuan Ni, Kai Chen, Kuan Cao, Liang Peng, Lin Qu, Minggang Wu, Peng Wang, Shuting Yu, Tingkun Wen, Wensen Feng, Xiaoxiao Xu, Yi~Wang, Yichang Zhang, Yongqiang Zhu, Yujia Wu, Yuxuan Cai, and Zenan Liu.
\newblock {Qwen-Image} technical report.
\newblock \emph{arXiv preprint arXiv:2508.02324}, 2025.

\bibitem[Xia et~al.(2025)Xia, Peng, Zhang, Huang, Liu, Li, Tan, Wu, Wang, Wang, Wu, Yu, and Jia]{exp_dreamomni2_2025}
Bin Xia, Bohao Peng, Yuechen Zhang, Junjia Huang, Jiyang Liu, Jingyao Li, Haoru Tan, Sitong Wu, Chengyao Wang, Yitong Wang, Xinglong Wu, Bei Yu, and Jiaya Jia.
\newblock {DreamOmni2}: Multimodal instruction-based editing and generation.
\newblock \emph{arXiv preprint arXiv:2510.06679}, 2025.

\bibitem[Yang et~al.(2026)Yang, Xu, Xie, Yang, Li, Lin, Zhu, Chen, Duan, Yue, Lin, Wang, and Pang]{exp_mmsi2026}
Sihan Yang, Runsen Xu, Yiman Xie, Sizhe Yang, Mo~Li, Jingli Lin, Chenming Zhu, Xiaochen Chen, Haodong Duan, Xiangyu Yue, Dahua Lin, Tai Wang, and Jiangmiao Pang.
\newblock {MMSI-Bench}: A benchmark for multi-image spatial intelligence.
\newblock In \emph{International Conference on Learning Representations}, 2026.

\bibitem[Ye et~al.(2023)Ye, Zhang, Liu, Han, and Yang]{ipadapter2023}
Hu~Ye, Jun Zhang, Sibo Liu, Xiao Han, and Wei Yang.
\newblock Ip-adapter: Text compatible image prompt adapter for text-to-image diffusion models.
\newblock \emph{arXiv preprint arXiv:2308.06721}, 2023.

\bibitem[Yeshwanth et~al.(2023)Yeshwanth, Liu, Nie{\ss}ner, and Dai]{scannetpp2023}
Chandan Yeshwanth, Yueh-Cheng Liu, Matthias Nie{\ss}ner, and Angela Dai.
\newblock Scannet++: A high-fidelity dataset of 3d indoor scenes.
\newblock \emph{arXiv preprint arXiv:2308.11417}, 2023.

\bibitem[Zhang et~al.(2026)Zhang, Wang, Liu, Jiang, Jiang, Sun, Yang, Zheng, Song, Huang, Duan, and Li]{twnv2026}
Yanbing Zhang, Bo~Wang, Jianhui Liu, Nan Jiang, Jiaxiu Jiang, Haoze Sun, Yijun Yang, Shenghe Zheng, Lin Song, Haoyang Huang, Nan Duan, and Wenbo Li.
\newblock {Thinking with Novel Views: A Systematic Analysis of Generative-Augmented Spatial Intelligence}.
\newblock \emph{arXiv preprint arXiv:2605.10588}, 2026.

\bibitem[Zhu et~al.(2026{\natexlab{a}})Zhu, Lin, Long, Cao, Wang, Pang, and Liu]{astra2026}
Chenming Zhu, Jingli Lin, Yilin Long, Peizhou Cao, Tai Wang, Jiangmiao Pang, and Xihui Liu.
\newblock Thinking with imagination: Agentic visual spatial reasoning with world simulators.
\newblock \emph{arXiv preprint arXiv:2606.06476}, 2026{\natexlab{a}}.

\bibitem[Zhu et~al.(2026{\natexlab{b}})Zhu, Jiang, Zheng, Luo, Zhong, Li, Wang, Rong, Liu, Chen, Lin, and Shen]{gsibench2026}
Muzhi Zhu, Shunyao Jiang, Huanyi Zheng, Zekai Luo, Hao Zhong, Anzhou Li, Kaijun Wang, Jintao Rong, Yang Liu, Hao Chen, Tao Lin, and Chunhua Shen.
\newblock Exploring spatial intelligence from a generative perspective.
\newblock \emph{arXiv preprint arXiv:2604.20570}, 2026{\natexlab{b}}.

\end{thebibliography}
